\documentclass[10pt,twocolumn,letterpaper]{article}

\usepackage{wacv}
\usepackage{times}
\usepackage{epsfig}
\usepackage{graphicx}
\usepackage{amsmath}
\usepackage{amssymb}
\usepackage{multirow}
\usepackage{mathtools}
\usepackage{subcaption}
\usepackage{comment}
\usepackage{pdfpages}
\usepackage{diagbox}

\usepackage[pagebackref=true,breaklinks=true,colorlinks,bookmarks=false]{hyperref}
\usepackage{silence}
\wacvfinalcopy 

\def\wacvPaperID{734} 

\DeclarePairedDelimiter\floor{\lfloor}{\rfloor}

\ifwacvfinal\fi
\begin{document}

\title{Geometric Image Correspondence Verification by Dense Pixel Matching}

\author{Zakaria Laskar\thanks{Equal contribution: \tt\small firstname.lastname@aalto.fi}\\
\and 
Iaroslav Melekhov\footnotemark[1]
\and
Hamed R. Tavakoli$^{2}$
\and 
Juha Ylioinas
\and
Juho Kannala \\
\small Aalto University, Espoo, Finland \hspace{1pt}
$^2$Nokia Technologies, Espoo, Finland
}

\maketitle
\thispagestyle{empty}

\begin{abstract}
This paper addresses the problem of determining dense pixel correspondences between two images and its application to geometric correspondence verification in image retrieval. The main contribution is a geometric correspondence verification approach for re-ranking a shortlist of retrieved database images based on their dense pair-wise matching with the query image at a pixel level. We determine a set of cyclically consistent dense pixel matches between the pair of images and evaluate local similarity of matched pixels using neural network based image descriptors. Final re-ranking is based on a novel similarity function, which fuses the local similarity metric with a global similarity metric and a geometric consistency measure computed for the matched pixels. For dense matching our approach utilizes a modified version of a recently proposed dense geometric correspondence network (DGC-Net), which we also improve by optimizing the architecture. The proposed model and similarity metric compare favourably to the state-of-the-art image retrieval methods. In addition, we apply our method to the problem of long-term visual localization demonstrating promising results and generalization across datasets.
\end{abstract}

\begin{figure}[t]
  \centering
  \includegraphics[width=1\linewidth]{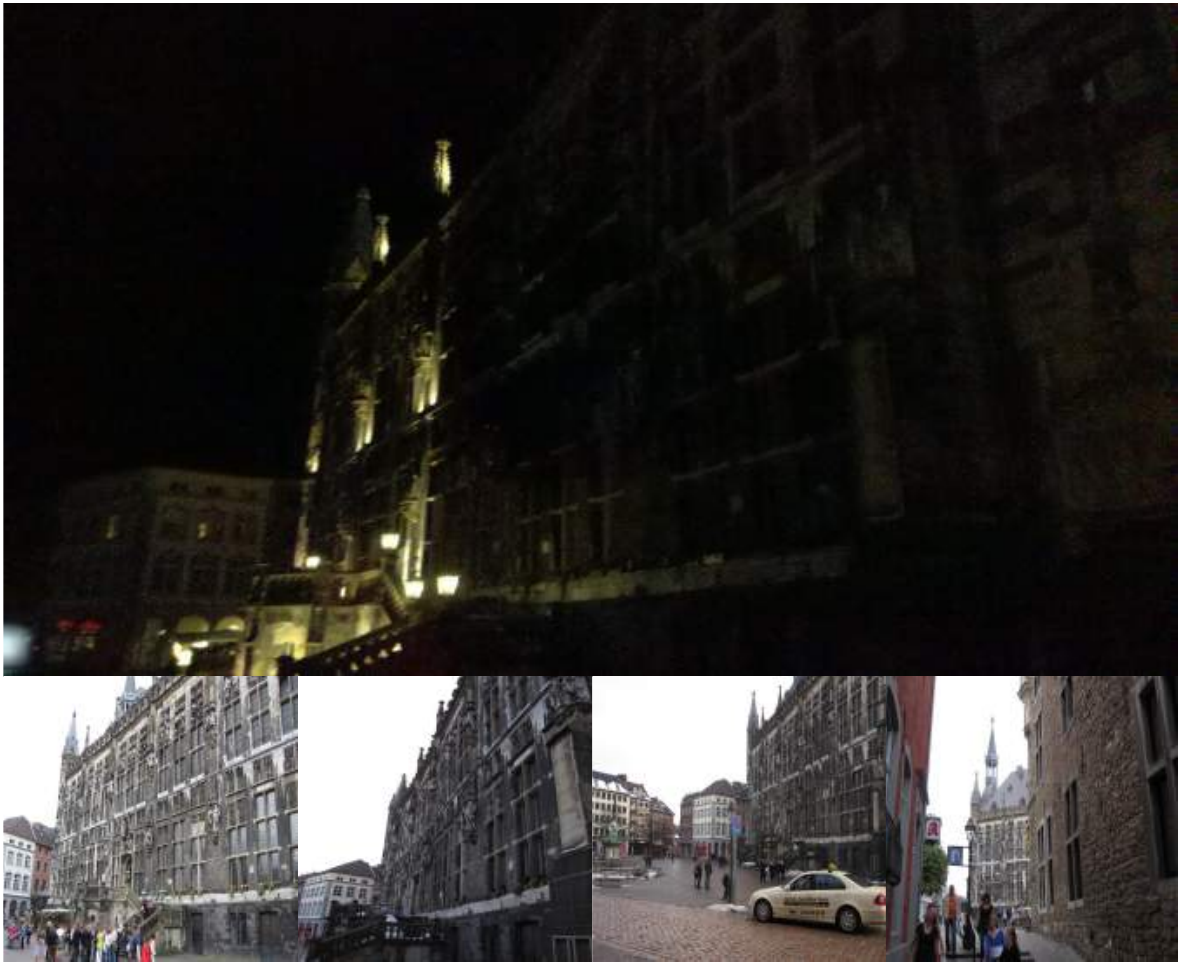}%
  \caption{Qualitative results of the proposed method for the task of image retrieval. The first row is a query taken at night-time with a mobile phone camera and the last row is a list of top-4 retrieved database images obtained by our method. All 4 are correct matches.
  }
  \label{fig:disclamer}
\end{figure}

\vspace{-2mm}
\section{Introduction}

Image retrieval is a well studied problem in the field of computer vision and robotics with applications in place recognition \cite{chen2011city, gronat2013learning, sattler2012image}, localization \cite{LaskarICCVW17,sattler2012image,taira2018inloc}, autonomous driving \cite{mcmanus2014shady}, and virtual reality~\cite{middelberg2014scalable} among many others. Given a query image, the image retrieval pipeline returns a ranked list of database images according to its measure of relevance to the query image. As raw pixels are not a good representation, extensive research has gone into finding discriminative and efficient image representations. The seminal work of Sivic and Zisserman~\cite{BoW} proposed Bag-of-Words based image representation using SIFT~\cite{Lowe2004}. Later, more advanced and efficient representations were proposed in the form of VLAD~\cite{vlad} descriptors and Fisher vectors~\cite{FisherVectors}. More recently, off-the-shelf~\cite{BabenkoNeuralCodes, Crow, ZakariaSCIA} and fine-tuned \cite{Netvlad, GordoECCV, Rad_eccv} convolutional neural network (CNN) representations have demonstrated great success in image retrieval. The models encode an input image to a global vector representation which leads to efficient retrieval allowing to use just a dot product as a similarity measure to obtain relevant database images. Once fine-tuned on auxiliary datasets with similar distribution as the target one, those methods have achieved state-of-the-art image retrieval performance \cite{Netvlad,GordoECCV,Rad_eccv}. However, the main limitation of such fine-tuned CNN representations are their generalization capabilities which is crucial in the context of city-scale localization where the database images can be quite similar in structure and appearance. Moreover, variations in illumination (\eg night time queries) or occlusion can significantly affect the encoded global representations degrading retrieval performance due to lack of spatial information. 

In this paper we leverage the advances of spatial geometry to obtain better ranking of the database images. To this end, we revisit the geometric verification problem in the context of image retrieval. That is, given an initial ranked list, $L$ of database images returned by a CNN model (\eg NetVLAD), we seek to re-rank a shortlist $L^{\prime} \in L$ of images by using dense pixel correspondences~\cite{DGCnet} which are verified by the proposed similarity functions. Previously, DGC-Net~\cite{DGCnet} has been successfully applied only to positive image pairs~\ie pairs with overlapping field of view. In this work we extend its applicability to verify positive and negative image pairs in the framework of geometric verification. That is, we demonstrate how dense pixel correspondence methods such as DGC-Net can be used to improve image retrieval by geometric verification.




In summary, the contributions of this work are threefold. First, we improve the baseline DGC-Net by constraining the matching layer to be locally and globally consistent. Second, we replace multiple decoders of the original DGC-Net architecture by the proposed universal decoder, 
 which can be shared for feature maps in different layers of the feature pyramid of DGC-Net. 
Third, we formulate two similarity functions, which first rank the shortlisted database images 
based on structural similarity and then re-rank them using appearance based similarity.

\begin{figure*}[t]
  \centering
    \includegraphics[width=.9\linewidth]{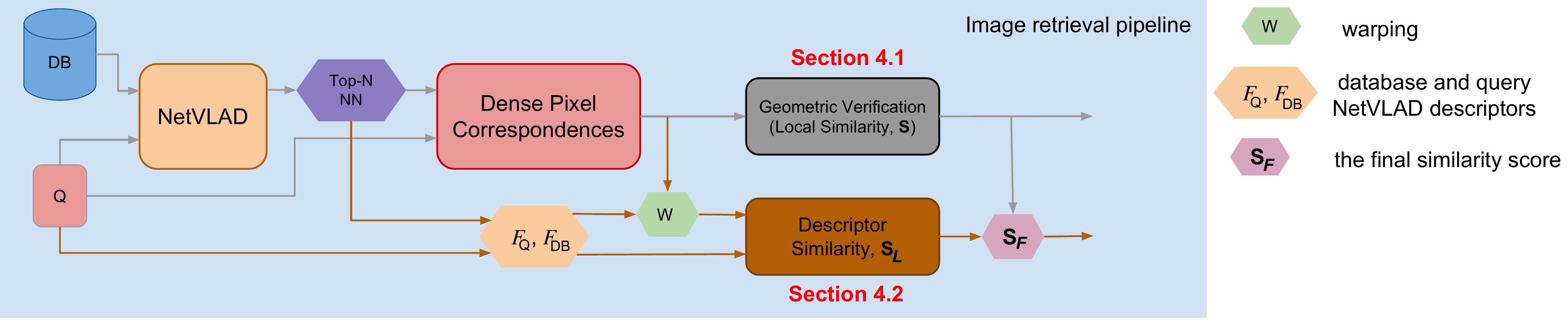}
  \caption{Overview of the proposed pipeline. Given a query image, we first rank the database images based on global similarity (\eg using NetVLAD). In the next stage dense pixel correspondences are computed between the query and top $N$ ranked database images. These correspondences are then verified by the proposed similarity functions utilizing geometry and CNN based image descriptors to re-rank database images according to the input query. See Sec.~\ref{sec:geometric_verification} and~\ref{sec:pixel_correspondence_estimation} for more details.}  
  
 \label{fig:fullpipeline}
 \vspace{-2mm}
\end{figure*}

\vspace{-2mm}
\section{Related work}
This work is closely related to image retrieval and image matching tasks. We provide a brief overview of existing approaches below.

\noindent\textbf{Image retrieval} methods can be broadly categorized into two categories: local descriptors~\cite{Chum2007TotalRA, JegouCVPR10, JegouCVPR07, MakadiaECCV10, BoW} and global representations~\cite{Netvlad,GordoECCV,Rad_eccv}. The approaches of the first category are based on either hand-engineered features such as SIFT~\cite{Lowe2004} or learnt CNNs descriptors on the task of local image patch matching~\cite{MelekhovPatchMatching, ZagoruykoPatchMatching}. Similarly, global representations methods can be further categorized into traditional hand-designed descriptors such as VLAD~\cite{vlad}, Fisher Vectors~\cite{FisherVectors}, Bag-of-Words~\cite{BoW} and CNN based methods \cite{Netvlad, BabenkoNeuralCodes, GordoECCV, Rad_eccv}. Babenko~\etal~\cite{BabenkoNeuralCodes} demonstrate that the performance of off-the-shelf CNN models pre-trained on ImageNet~\cite{Imagenet} fall behind traditional local descriptors. However, when trained on an auxiliary dataset, the performance improves over such hand-engineered descriptors~\cite{Netvlad, GordoECCV, Rad_eccv}.

In addition to the standard retrieval approaches, there are several methods that attempt to explain the similarity between the query and top ranked database images using a geometric model~\cite{ChumCVPR05, MatasICCV05, taira2018inloc}. The geometric model is estimated by fitting a simple transformation model (\eg planar homography) to the correspondence set obtained using local descriptors such as SIFT, or off-the-shelf CNN descriptors~\cite{taira2018inloc}. In this work, we also use pre-trained CNN descriptors. However, in contrast to~\cite{taira2018inloc} which uses exhaustive nearest-neighbor search in descriptor space, we model the similarity using a learnt convolutional decoder. Moreover,~\cite{taira2018inloc} only uses coarse correspondence estimate, while our similarity decoder allows fine high resolution pixel level correspondence estimation. This is particularly important in city scale localization due to subtle differences in an overall similar architectural style observed in this scenario  (\cf Fig.~\ref{fig:quality_retrieval_tokyo_aachen}). 




\noindent\textbf{Image matching.} This task relates to the optical flow estimation problem. Recently proposed optical flow methods~\cite{FlowNet2, PWC-Net} utilize a local correlation layer that performs spatially constrained matching in a coarse-to-fine manner. DGC-Net~\cite{DGCnet} extends this process of learning iterative refinement of pixel correspondences using a global correlation layer to handle wide viewpoint changes in the task of instance matching. Such a global correlation layer for instance matching has been used to estimate geometric transformations~\cite{Rocco17}. Melekhov~\etal~\cite{DGCnet} demonstrate that such a method falls behind dense correspondence approaches due to the constrained range of transformations estimated by~\cite{Rocco17}. Recently, Rocco~\etal~\cite{RoccoNeigh} propose locally and globally constrained matching network on top of the global correlation layer which leads to improvement in instance and semantic matching. However, such a global correlation layer can only provide coarse correspondence estimates.



\vspace{-2mm}
\section{Method overview}\label{sec:method}

Our contributions are related to the two last stages of the following three-stage image retrieval pipeline: \emph{1)} Given a query image, we retrieve a shortlist of relevant database images using a fast and scalable retrieval method based on representing images with a descriptor vector; \emph{2)} We perform dense pixel matching between the query and each shortlisted database image in a pairwise manner using a correspondence estimation network; \emph{3)} We determine a set of cyclically consistent dense pixel matches for each image pair and use them to compute a similarity metric, which provides the final re-ranking of the shortlist.

The particular architecture of the aforementioned retrieval pipeline used in this work is illustrated in Fig.~\ref{fig:fullpipeline}. That is, we use NetVLAD \cite{Netvlad} for the first stage, our own modified version of DGC-Net \cite{DGCnet} for the second stage, and the proposed approach with a novel similarity metric for the third stage. Here NetVLAD is used for retrieval, but also other global image level descriptors could be used instead.

Our contributions related to stages \emph{2)} and \emph{3)} above are described in the following sections. The geometric verification method is presented in Section~\ref{sec:geometric_verification} and our modifications to the DGC-Net architecture are described in Section~\ref{sec:pixel_correspondence_estimation}.

\vspace{-2mm}
\section{Geometric verification}\label{sec:geometric_verification}
Dense pixel correspondences produced by~\cite{DGCnet} do not take into account the underlying model explaining the 3D structure of the scene by the image pair. 
RANSAC~\cite{RANSAC} has been a popular method of choice to find the set of inliers from the whole correspondence set. However, dense pixel correspondences predicted by CNNs~\cite{DGCnet} are locally smooth due to the shared convolutional filters at different layers. As a result, RANSAC usually finds a large set of inliers even for non-matching image pairs. We propose two methods to eliminate these limitations in the following sections. That is,  given an initial ranked shortlist $L$ of database images based on global representation similarity with the query image, we re-rank a new shortlist $L' \subseteq {L}$ through a series of geometric verification steps (Sec.~\ref{sec:cyc} and \ref{sec:global_and_local}). 

\subsection{Cyclically consistent geometry}
\label{sec:cyc}



We propose a similarity cost function, $S$ that combines RANSAC based geometric model estimation with cyclic consistency. Given a dense pixel correspondence map, $O \in \mathbb{R}^{H \times W  \times 2}$, RANSAC outputs a set of inliers, $I \subseteq O$~\wrt to a transformation model (\eg planar homography). We then estimate the subset of inliers that are cyclically consistent, $C \subseteq I$ using forward and backward correspondence maps predicted by our network (\ie~$O^A$ and $O^B$). The cyclically consistent matches are those matches for which the combined mapping  $O^A\! \circ \! O^B$ is close to an identity mapping. For geometrically dissimilar images, cyclic consistency constraint further constrains the number of inliers as the assumption here is that transformation model obtained by RANSAC may be inconsistent in forward and backward directions. We define this similarity function as follows
\begin{equation}
S = \frac{|C|}{|I|} \cdot exp\left(-\frac{\beta}{|C|}\right),
\label{eq:s_sim}
\end{equation}
\noindent where $\beta$ is a constant. As $|C|/|I|$ is a ratio, the exponential term is added to down-weight the similarity cost for image pairs which have less cyclically consistent correspondences in the inlier set. As $\beta$ must be greater than $|C|$, we set it to 240x240 which is the maximum value of $|C|$ as our dense correspondence network (Sec. \ref{sec:pixel_correspondence_estimation}) operates on fixed size images of resolution 240x240. The similarity is computed in both directions, $S^A,S^B$ and the final similarity is the maximum of the two values, $S = max(S^A,S^B)$. The shortlist $L$ is re-ranked using $S$ resulting in the new shortlist $\hat{L}$.

\begin{figure*}[t!]
  \centering
    \includegraphics[width=.9\linewidth]{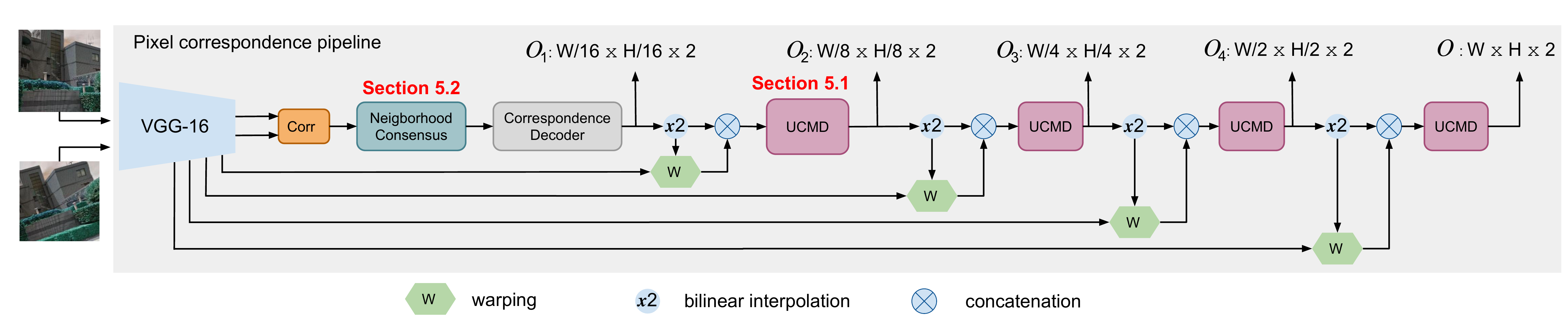}
  \caption{Overview of the dense pixel correspondence network. Pre-trained VGG-16 network is used to create a multiscale feature pyramid $P$ of the input image pair. Correlation and neighborhood consensus layers use features from the top level of $P$ to establish coarse pixel correspondences which are then refined by the proposed unified correspondence map decoder (UCMD). In contrast to DGC-Net~\cite{DGCnet} with multiple decoders, UCMD can be applied to each level of the multi-scale feature pyramid seamlessly leading to smaller memory footprint.}
 \label{fig:f_pipeline}
 \vspace{-2mm}
\end{figure*}

\subsection{Global and local similarity}
\label{sec:global_and_local}
Using the geometry based similarity function $S$ to re-rank the shortlist typically improves retrieval accuracy, but the retrieved list may still contain outliers as the global and local appearance similarity is not directly taken into account while computing $S$. Hence, the top-ranked database images in the geometrically verified shortlist $\hat{L}$ are passed through a second similarity function based on global and local descriptor similarity. The second similarity function is detailed below and more costly to evaluate, as it requires dense image feature extraction on high resolution images (\eg 640x480 or higher) to obtain high resolution feature maps. On the other hand the dense correspondence estimation in Eq. \ref{eq:s_sim} is performed on lower resolution images (240x240) and hence is significantly faster to compute. Therefore we have a two-stage re-ranking, where the second re-ranking is done only for a subset of top-ranked images from the first stage.

To obtain global dissimilarity $G$ we use normalized global descriptors from a pre-trained network NetVLAD \cite{Netvlad}. The network was originally trained to learn powerful representations for image retrieval. The Euclidean distance between the global representations is defined as the global dissimilarity value $G$. To compute local similarity, we extract hypercolumn \cite{RossCVPR15} features from different NetVLAD layers (see Supplementary), L2 normalize and concatenate them along channel dimension. The final features are again L2 normalized resulting in feature maps, $\tilde{F}_A, \tilde{F}_B$, where $\tilde{F} \in \mathbb{R}^{H \times W \times Z}$, and $(H,W), Z$ are the image resolution and the final descriptor length. The local descriptor similarity $S_L$ is then obtained as: 
\begin{equation}
    S_L = \sum_{a} \left(f_{A}^a \cdot f_{B}^a\right) m^a
\label{eq:descp_sim}
\end{equation}
\noindent where $ \cdot $ denotes inner product, $f_{A}^a \in {}^{w}\tilde{F}_{A}$ and $f_{B}^a \in \tilde{F}_{B}$ are the hypercolumn NetVLAD features at location $a$ in the warped source $\tilde{F}_A$ and target feature map $\tilde{F}_B$, and $m^a \in M$, where $M$ is the mask containing \textbf{1}s at cyclically consistent pixels. Thus, Eq.~\ref{eq:descp_sim} computes the cosine similarity between normalized warped source and target hypercolumn descriptors at cyclically consistent pixel locations.

The final similarity function between an image pair is a function of global dissimilarity and local similarities, $G$ and $S_L$ : 
\begin{equation} \label{eq:final_sim}
    S_F = \log_{10}\left(S_L \cdot S\right) \cdot 10^{-G}
\end{equation}
Here, local similarity score $S_L$ is weighted by the similarity score $S$. We use $S_F$ to re-rank the top-ranked images in $\hat{L}$ to get the final shortlist $L'$ for a given query. The $log$ term is added as a normalization to balance the local and global scores. Although there are many possible ways to combine the local and global scores, we perform an extensive evaluation (see Supplementary) and show that the current form of these equations (\ref{eq:s_sim} and \ref{eq:final_sim}) achieves the best performance. 


\section{Pixel correspondence estimation}~\label{sec:pixel_correspondence_estimation}

To obtain dense matching between two images we use a CNN network based on the architecture of DGC-net proposed by~\cite{DGCnet}. In this section, we provide two modifications to DGC-Net leading to more compact but effective model.



\begin{figure}[t]
  \centering
    \includegraphics[width=.9\linewidth]{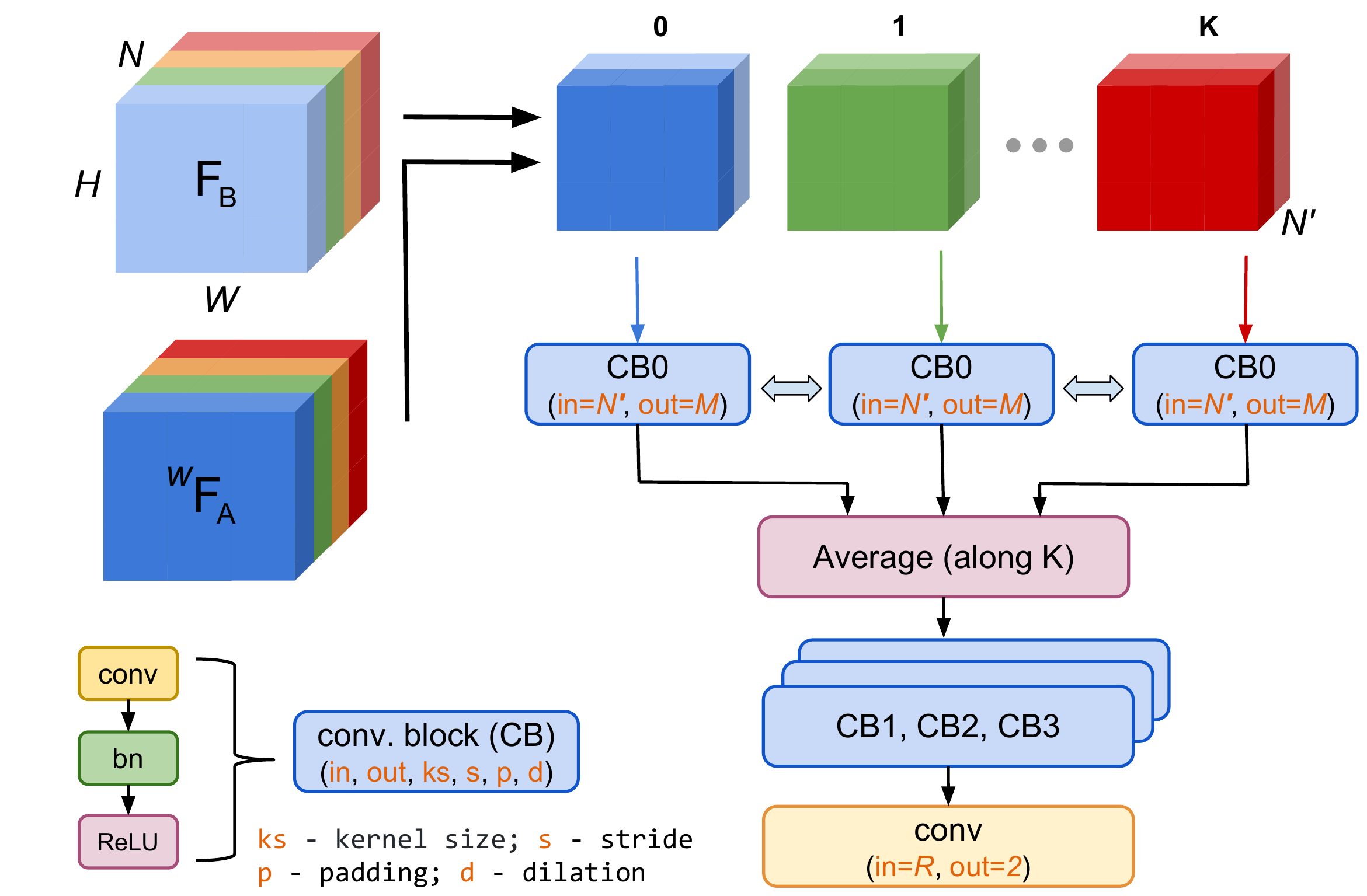}
  \caption{Overview of the unified correspondence map decoder (UCMD) $D_c$. The feature maps of the target $F_{B}$ and the warped source ${}^{w}F_{A}$ images have been split into $k$ tensors and then concatenated along the channel dimension. Further, each tensor is complemented by the correspondence map estimates $H \times W \times 2$ (expelled from the figure for clarity) and then fed into a convolutional block $CB0$ with $N^{\prime}$ inputs and shared weights. The output feature maps of $CB0$ are then averaged and processed by the remaining layers of the decoder to produce refined pixel correspondence estimates.}
 \label{fig:f_channel_convolution}
 \vspace{-2mm}
\end{figure}

\subsection{Unified correspondence map decoder}\label{ssec:channel_conv}

In general, DGC-Net consists of a series of convolutional layers and activation functions as an encoder $E$ with $M$ layers. An input image pair $I_A, I_B \in \mathbb{R}^{H \times W \times 3}$ is fed into the encoder independently to obtain a multi-resolution feature pyramid, $P = \{(F_{A}^l, F_{B}^l)|l=1,2,...M\}$. Here $F^l \in \mathbb{R}^{H_l \times W_l  \times N_l}$ is the feature map at the output of layer $l$ of the encoder. The encoded feature maps at the top level of $P$, $(F_{A}^M, F_{B}^M)$ are passed through a global correlation layer, that computes exhaustive pairwise features cosine similarity. The output of correlation layer is then passed through a decoder $D_1$ that estimates the initial correspondence map $O_1$ at the same resolution as $F^M$. $O_1$ is then iteratively refined by a series of decoders $D = \{D_2,D_3,... D_M\}$ to obtain the final correspondence grid $O_M$ at the same resolution as input images. Each decoder, $D_j \in D$ takes in as input $X_j = \{O_{j-1}, {}^{w}F_{A}^{M-j+1}, F_{B}^{M-j+1}\}$, where $O_{j-1}$ is the upsampled correspondence map estimated by the previous decoder $D_{j-1}$, ${}^{w}F_{A}^{M-j+1}$ and $F_{B}^{M-j+1}$ are the warped source and target feature maps at $l=M-j+1$. However, since feature maps at level $l$ of $P$ have various number of channels, each decoder $D_l$ has different structure which leads to increased memory costs.

In this work, we propose a unified correspondence map decoder $D_c$ (UCMD) illustrated in Fig.~\ref{fig:f_channel_convolution}. The unified decoder behaves like a recursive refinement function that operates on feature maps across different layers $l$ of $P$. 
More specifically, we divide the concatenated input feature maps in $X_j$ into $k_j$ non-overlapping components as shown in Fig.~\ref{fig:f_channel_convolution}. We then propagate each of the $k_j$ concatenated components $X_j^t, t=1,..k_j$ through the first convolutional layer (CB0) of our decoder, $D_c$. The resulting $k_j$ feature maps at the output of CB0 are subsequently averaged and passed through the remaining layers to obtain refined correspondence estimates $O_j$. 


The number of inputs of CB0 is $N^{\prime}=2Q+2$, where $Q$ specifies the number of channels in feature maps ${}^{w}F_{A}, F_{B}$ which are concatenated along the channel dimension. The additional 2 channels comprise of the upsampled coarser pixel correspondence map estimate from the previous layer of $P$. Therefore, $k_l$ is given by $\floor{N_l/Q}$ where $N_l$ is the dimensionality of the feature maps at the current layer $l$.  

\noindent\textbf{Inference.} During the testing phase, apart from evaluating the trained network directly we additionally follow a second strategy. We infer the pixel correspondences by feed-forwarding each $X_j^t$ through the complete decoder $D_c$ resulting in $k$ correspondence map estimates $O^k$. The process is applied to each level of the feature pyramid $P$. The mean $\mathbb{E}(O^k)$ is used as the final pixel correspondence map estimate. This formulation was not used during training as it did not lead to convergence.

\subsection{Match consistency}\label{ssec:match_consistency}
\label{ncnet}
The global correlation layer only measures the similarities in one direction~\ie from target to source image. However, many related works in the optical flow have shown that cyclic consistency allows the network to achieve better performance. In \cite{RoccoNeigh}, a similar kind of global correlation layer was applied with cyclic consistency and neighborhood consensus to learn optimal feature correspondence. The idea is that matches should be consistent both locally and cyclically. That is nearby matches should be locally consistent and also the matches should be consistent in both forward and backward direction. Thereby, we integrated the Neighborhood Consensus Network (NCNet) \cite{RoccoNeigh} in our network. 
 In contrast to original DGC-Net, the output of the correlation layer is now passed through NCNet with learnable parameters before being feed-forwarded through the decoders $D_M$ and $D_c$ to obtain dense pixel correspondences $O$. We refer to this network as~\texttt{DGC-NC-UCMD-Net}.

\begin{figure*}[t!]
 	\centering
 	\begin{subfigure}[t]{.2\textwidth}
 		\centering
 		\includegraphics[width=\textwidth]{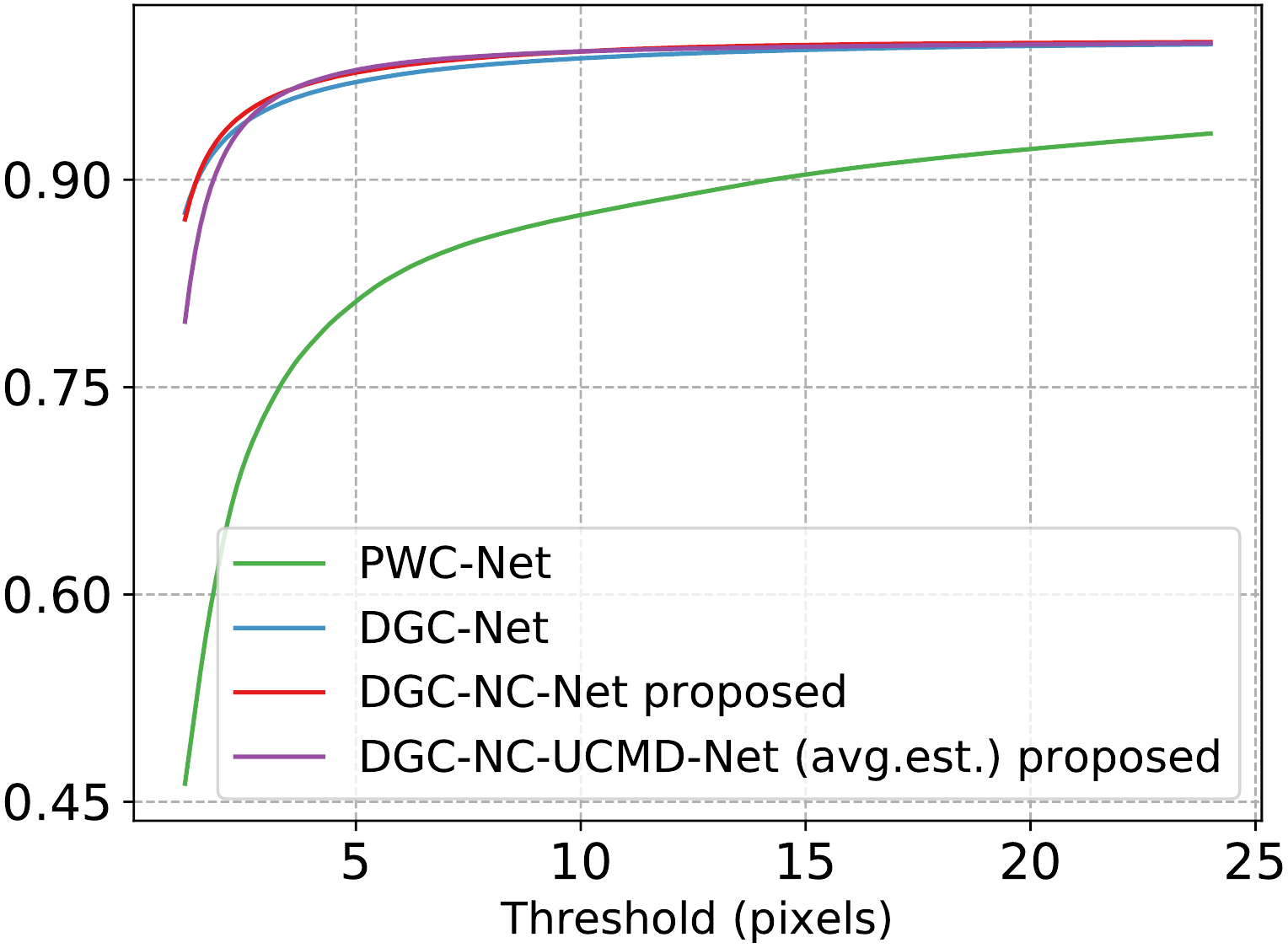}
 		\caption{Viewpoint I}
 	\end{subfigure}%
 	~
 	\begin{subfigure}[t]{.2\textwidth}
 		\centering
 		\includegraphics[width=\textwidth]{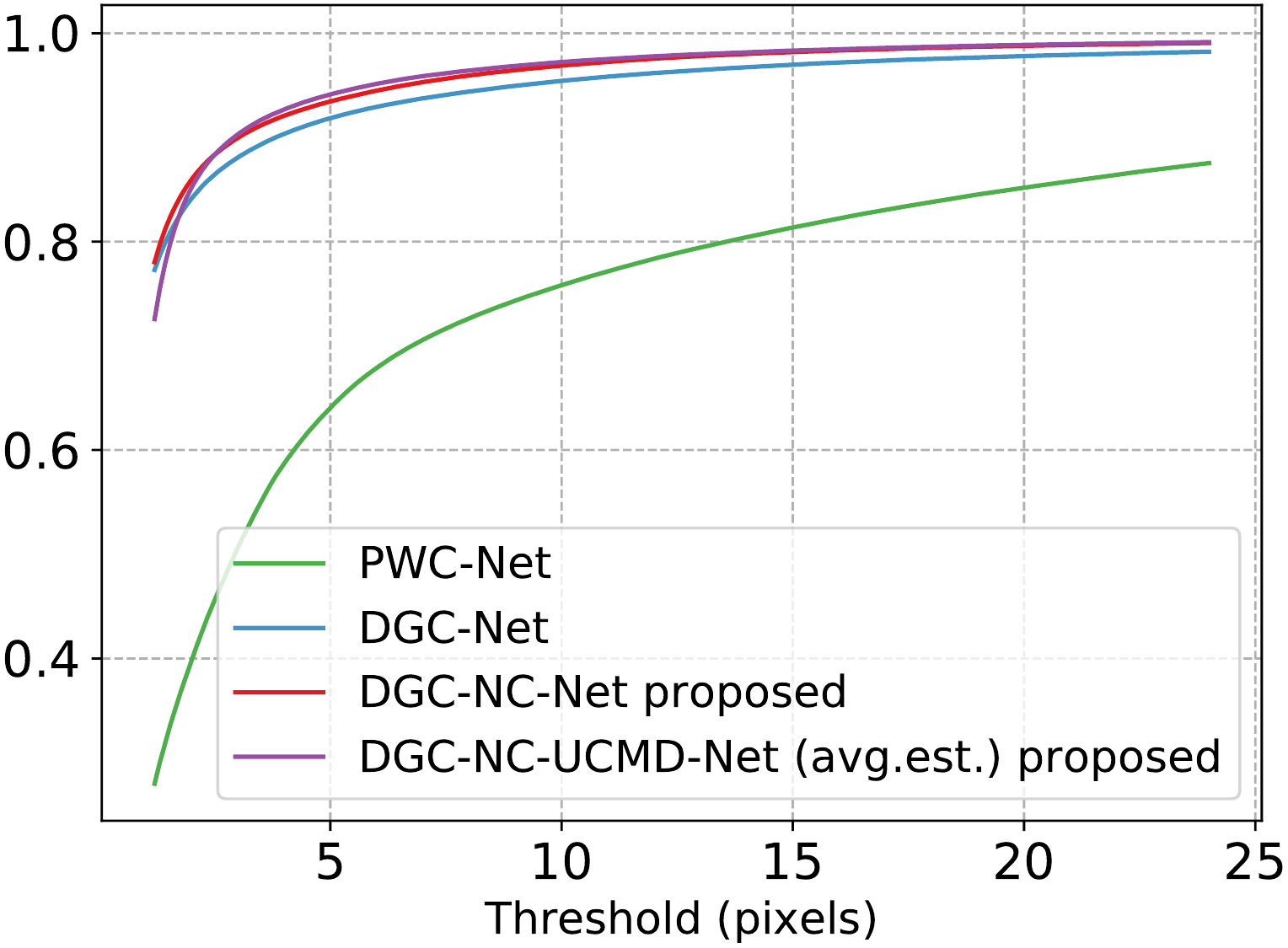}
 		\caption{Viewpoint II}
 	\end{subfigure}%
 	~
 	\begin{subfigure}[t]{.2\textwidth}
 		\centering
 		\includegraphics[width=\textwidth]{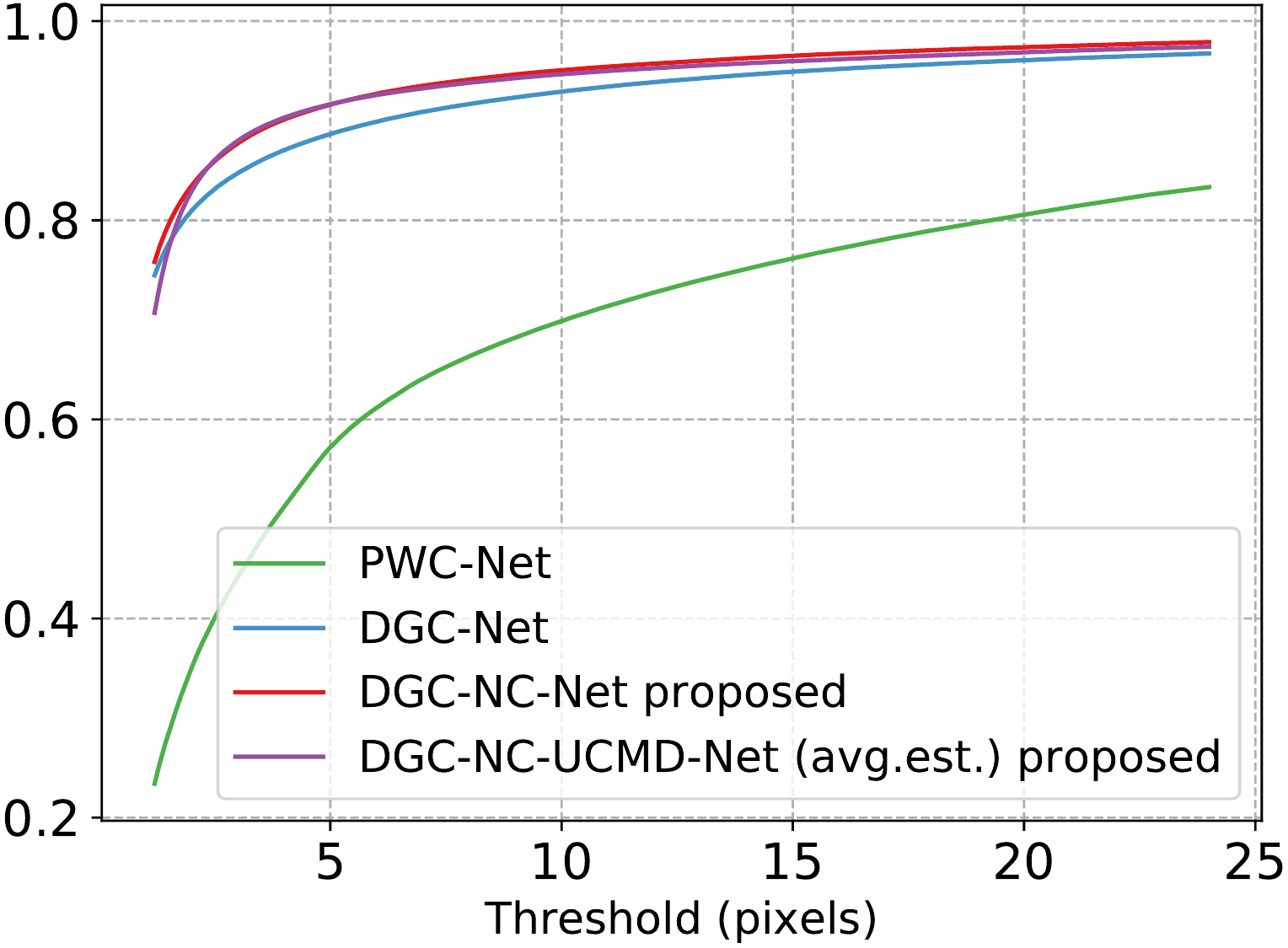}
 		\caption{Viewpoint III}
 	\end{subfigure}%
 	~
 	\begin{subfigure}[t]{.2\textwidth}
 		\centering
 		\includegraphics[width=\textwidth]{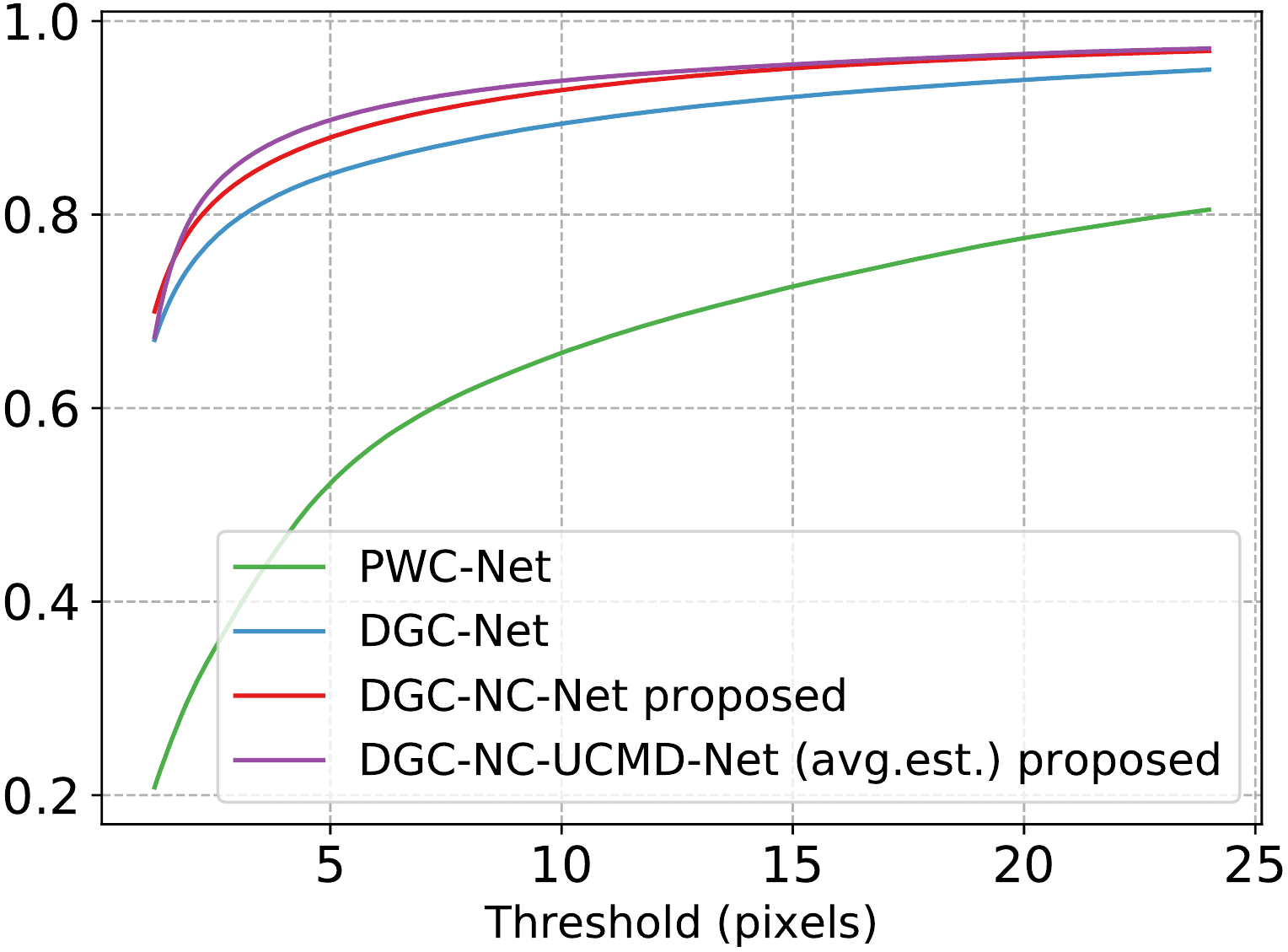}
 		\caption{Viewpoint IV}
 	\end{subfigure}%
 	~
 	\begin{subfigure}[t]{.2\textwidth}
 		\centering
 		\includegraphics[width=\textwidth]{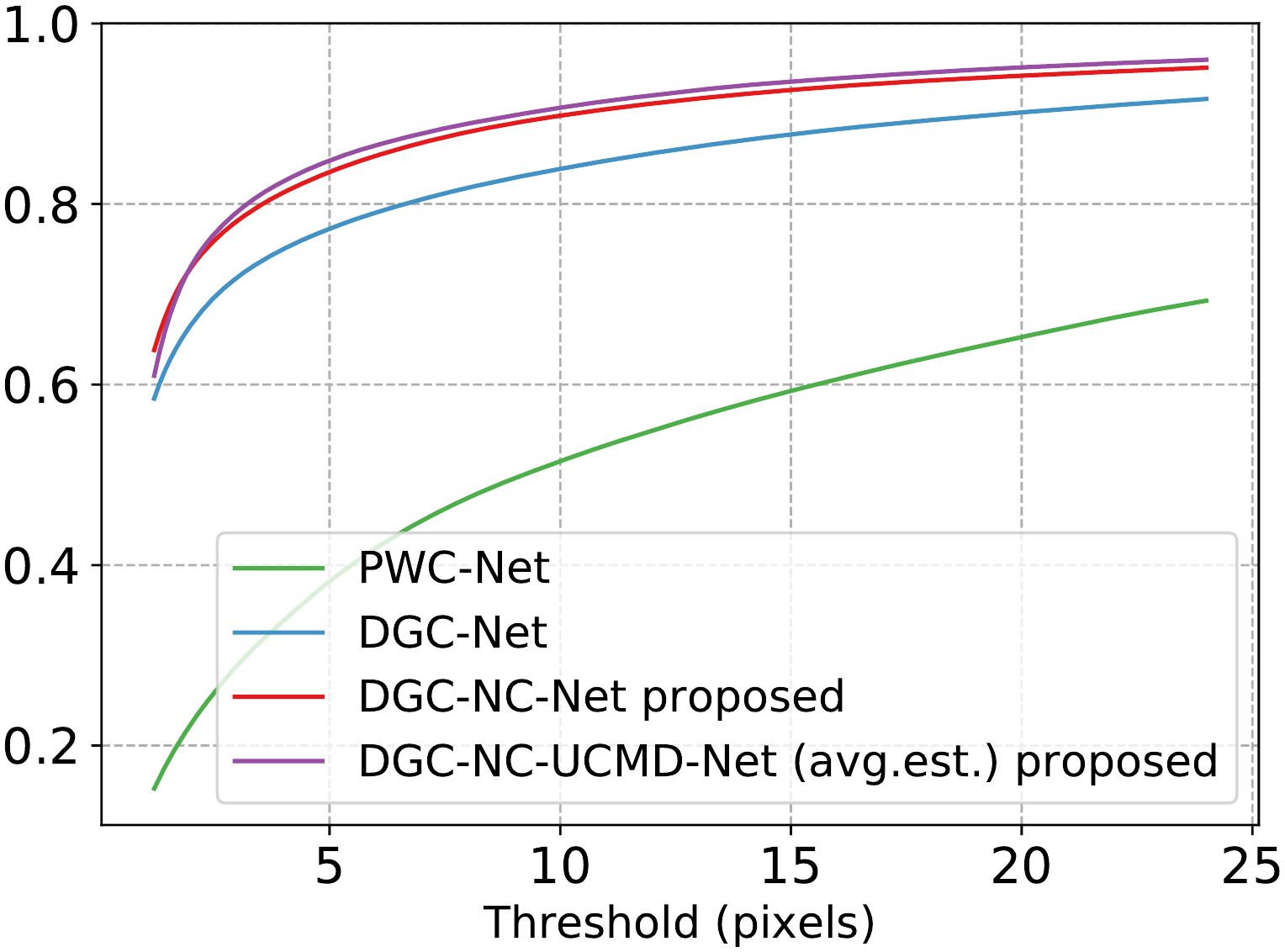}
 		\caption{Viewpoint V}%
 	\end{subfigure}
\caption{PCK metric calculated for different Viewpoint IDs of the HPatches dataset. The proposed architectures (DGC-NC-*) substantially outperform all strong baseline methods with a large margin.}\label{fig:pck_hpatches}
\end{figure*}

\section{Experiments}
We  discuss  the  experimental  settings  and  evaluate  the proposed method on two closely related tasks,~\ie establishing dense pixel correspondences between images (image matching) and retrieval-based localization.

\subsection{Image matching}
\label{sec:img_match_exp}
For this task we compare our approach with DGC-Net~\cite{DGCnet},  
 which can handle strong geometric transformations between two views. We use training and validation splits proposed by~\cite{DGCnet} to compare both approaches fairly. More specifically, diverse synthetic transformations (affine, TPS, and homography) have been applied to Tokyo Time Machine dataset~\cite{Netvlad} to generate ~20k training samples. Similarly to~\cite{DGCnet}, the proposed network has been trained by minimizing $L1$ distance between the ground-truth and estimated correspondence map $O_i$ at each level of the feature pyramid $P$ (\cf Fig.~\ref{fig:f_pipeline}). Details of the training procedure are given in supplementary.

We evaluate our method on HPatches dataset~\cite{HPatches} and report the average endpoint error (AEPE) of the predicted pixel correspondence map. HPatches dataset consists of several sequences of real images with varying photometric changes. Each image sequence represents a reference image and 5 corresponding source images taken under a different viewpoint with the estimated ground-truth homography $\mathbf{H}$. As predicting a dense pixel correspondence map is closely related to optical flow estimation, we provide AEPE for strong optical flow (OF) baseline methods, \ie FlowNet2~\cite{FlowNet2} and PWC-Net~\cite{PWC-Net} respectively.

\begin{table}[t!]
\begin{center}
\resizebox{.49\textwidth}{!}{%
    \begin{tabular}{l|l l l l l}
        \multirow{2}{*}{Method} & \multicolumn{5}{|c}{Viewpoint ID} \\
        & \multicolumn{1}{|c}{I} & \multicolumn{1}{c}{II} & \multicolumn{1}{c}{III} & \multicolumn{1}{c}{IV} & \multicolumn{1}{c}{V}\\
    \hline
    \hline
    FlowNet2~\cite{FlowNet2} & 5.99 & 15.55 & 17.09 & 22.13 & 30.68 \\
    PWC-Net~\cite{PWC-Net} & 4.43 & 11.44 & 15.47 & 20.17 & 28.30 \\
    Rocco~\cite{Rocco17} & 9.59 & 18.55 & 21.15 & 27.83 & 35.19 \\
    DGC-Net~\cite{DGCnet} & 1.55 & 5.53 & 8.98 & 11.66 & 16.70 \\
    \hline
    DGC-NC-UCMD-Net & \ 1.90 & 5.02 & 9.08 & 10.18 & 13.24 \\
    DGC-NC-UCMD-Net (avg. est.) & \ 1.51 & 4.46 & 8.66 & \textbf{9.59} & \textbf{12.62} \\
    DGC-NC-Net & \textbf{1.24} & \textbf{4.25} & \textbf{8.21} & 9.71 & 13.35 \\
    \hline
\end{tabular}
}
\end{center}
\vspace{-2mm}
\caption{AEPE metric for different viewpoint IDs of the HPatches dataset (lower is better). 
}\label{tbl:perf_hpatches}
\end{table}

We calculate AEPE over all image sequences belonging to the same Viewpoint ID of the HPatches dataset and report the results in Tab.~\ref{tbl:perf_hpatches}. Here, DGC-NC-Net refers to the original DGC-Net architecture complemented by NC layer (Sec.~\ref{ssec:match_consistency}) with a set of independent decoders at each level of the spatial feature pyramid $P$. Compared to DGC-Net, this model can achieve better performance reducing the overall EPE by 20\% for the most extreme viewpoint difference between the reference and source images (Viewpoint V). According to Tab.~\ref{tbl:perf_hpatches}, DGC-NC-UCMD-Net with one universal correspondence map decoder (Sec.~\ref{ssec:channel_conv}) falls slightly behind of DGC-NC-Net (by 12\% in average across all Viewpoint IDs) but it demonstrates advantages in terms of computation and memory costs (c.f. Sec. \ref{ssec:ablation} and Supplementary). However, DGC-NC-UCMD-Net performance can be improved further if, at inference time, rather than averaging $k$ feature maps produced by the first convolutional block of UCMD (\cf Fig.~\ref{fig:f_channel_convolution}) we average \textit{predicted pixel correspondence estimates} for each input $k$ feature map. We refer this model as DGC-NC-UCMD-Net (avg. est.). 

In addition, we report  a  number  of  correctly  matched pixels between two images by calculating PCK (Percentage of Correct Keypoints) metric with different thresholds. As shown in Fig.~\ref{fig:pck_hpatches}, the proposed DGC-NC-* models outperform DGC-Net by about 4\% and correctly match around 62\% pixels for the case where geometric transformations are the most challenging (Viewpoint V).

\subsection{Localization}
\label{sec:loc}
We study the performance of our pipeline in the context of image retrieval for image based city-scale localization. 
For evaluating the performance of our pipeline, we consider three localization datasets: Tokyo24/7~\cite{Tokyo24/7}, Aachen Day-Night~\cite{AachenDataset}, and extended CMU-Seasons~\cite{AachenDataset}. For all the datasets, we follow the same procedure outlined below. For a given query we first obtain a ranked list of database images, ${L}$ based on Euclidean distance between their global NetVLAD representations, $G$. The top 100 ranked database images, $\hat{L} \subseteq {L}$ are re-ranked according to their geometric similarity score based on $S$. From these geometrically verified re-ranked database images, we pass the top 20, $L' \subseteq \hat{L}$ through the more expensive and stricter representation similarity function, $S_F$. Based on this final similarity, the final re-ranking is done on $L'$.   

\noindent\textbf{Localization metrics.} The performance on the Tokyo24/7 dataset is evaluated using Recall@N, which is the number of queries that are correctly localized given $N$ nearest-neighbor database images returned by the model. The query is considered correctly localized if at least one of the relevant database images is presented in the top $N$ ranked database images. In contrast, the localization performance on Aachen Day-Night and extended CMU-Seasons is measured in terms of accuracy of the estimated query pose. The accuracy is defined as the percentage of queries with their estimated 6DOF pose lying within a pre-defined threshold to the ground-truth pose.

\begin{figure}[t!]
  \centering
    \includegraphics[width=.8\linewidth]{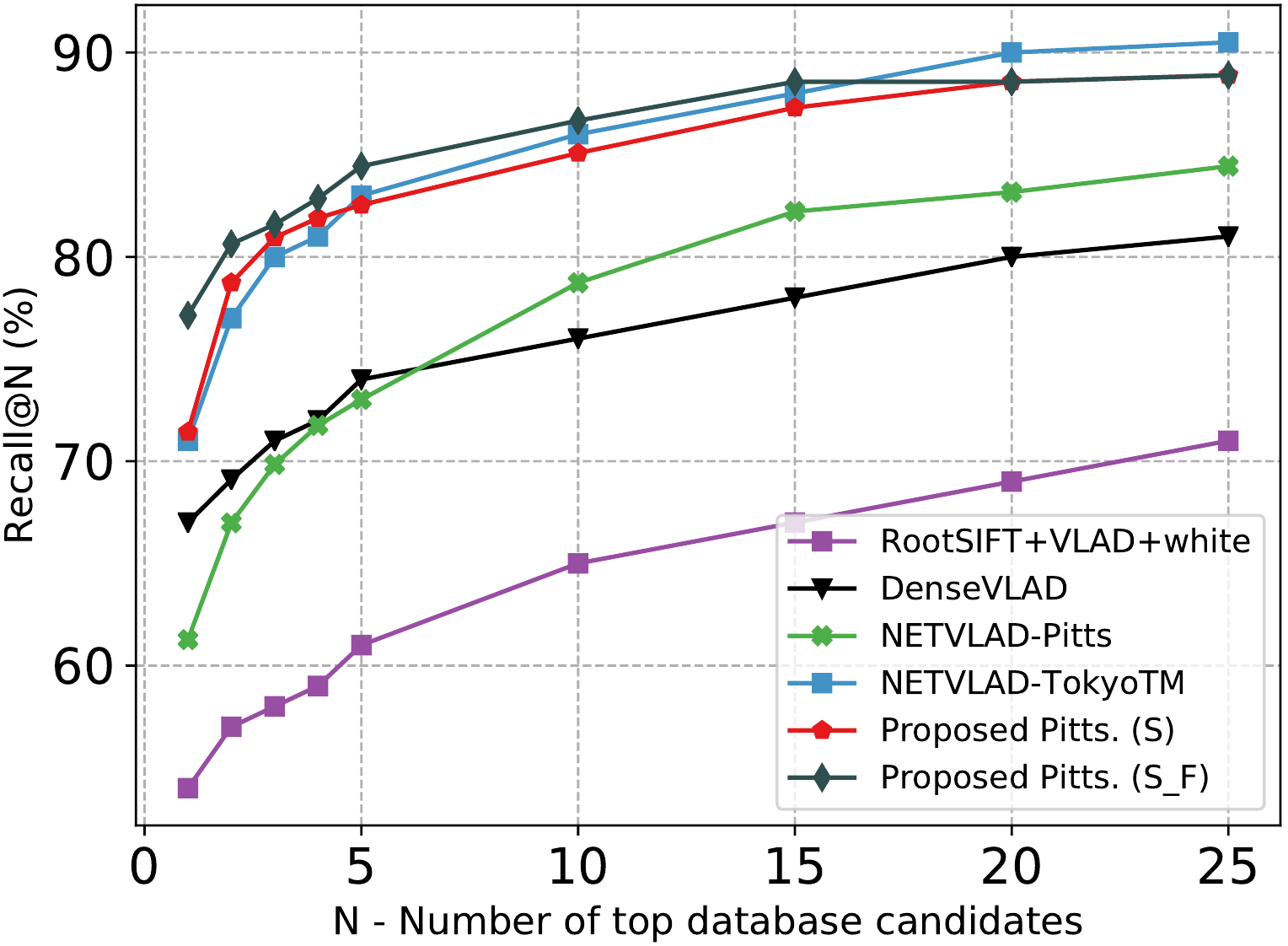}
  \caption{Comparison of the proposed methods versus state-of-the approaches for place recognition.}
 \label{fig:f_recall_res}
 \vspace{-2mm}
\end{figure}

\begin{table}[t!]
\begin{center}
\resizebox{.35\textwidth}{!}{%
    \begin{tabular}{l|l l l}
        \multirow{2}{*}{Methods} & \multicolumn{3}{|c}{Recall} \\
        & \multicolumn{1}{|c}{r@1} & \multicolumn{1}{c}{r@5} & \multicolumn{1}{c}{r@10}\\
    \hline
    \hline
    DenseVLAD~\cite{Tokyo24/7} & 67.1 & 74.2 & 76.1 \\
    NetVLAD-Pitts & 61.27 & 73.02 & 78.73 \\
    NetVLAD-TokyoTM~\cite{Netvlad} & 71.1 & 83.1 & 86.2 \\
    \hline 
    SIFT & 73.33 & 80.0 & 84.4 \\
    Inloc~\cite{taira2018inloc} & 62.54 & 67.62  & 70.48 \\
    Proposed $(S)$ Pitts & 71.43 & 82.54 & 85.08 \\
    Proposed $(S_F)$ Pitts & \textbf{77.14} & \textbf{84.44} & \textbf{86.67} \\
    \hline
\end{tabular}
}
\end{center}
\vspace{-2mm}
\caption{Localization performance on the Tokyo24/7 dataset (higher is better). Our proposed method outperforms Inloc and SIFT based geometric verification.}\label{tab:tokyo_loc}
\end{table}

\noindent\textbf{Tokyo 24/7.} We compare the proposed approach with several strong baseline methods for place recognition. The hand-crafted methods are represented by DenseVLAD~\cite{Tokyo24/7} which aggregates densely extracted SIFT descriptors~\cite{Lowe2004}. As our re-ranking method aims to improve the initial ranking by NetVLAD representations, we consider NetVLAD as a baseline. In particular, we use a publicly available PyTorch implementation of NetVLAD trained on Pittsburgh dataset (NetVLAD-Pitts). As a reference, NetVLAD-Pitts obtains $85.2/94.8/97.0\%$ compared to $84.1/94.6/95.5\%$ by NetVLAD \cite{Netvlad} on Pitts-30k \cite{Netvlad} validation set. In addition, we also consider Inloc \cite{taira2018inloc} which uses dense NetVLAD descriptors in a geometric verification setting to obtain the final shortlist of ranked database images. 

The Recall@N for the baseline methods are presented in Fig.~\ref{fig:f_recall_res}. Our geometric verification based pipeline achieves the state-of-the-art performance at Recall@1-10. 
The proposed approach significantly outperforms NetVLAD-Pitts and other baseline methods for all Recall@N thresholds (\cf Tab.~\ref{tab:tokyo_loc}). Moreover, it is noteworthy that our method pushes the generalization performance of NetVLAD-Pitts above the NetVLAD-TokyoTM which was trained on images with similar distribution as Tokyo24/7. We also compared against traditional SIFT \cite{RootSIFT} based geometric verification which achieved 73.33\% for Recall@1. We used COLMAP \cite{colmap_1,colmap_2} to extract SIFT features, followed by fundamental matrix based geometric verification to compute the inlier count.


\noindent\textbf{Aachen Day-Night and Extended CMU-Seasons.} Most localization systems involve an image retrieval stage where our proposed method can be directly applied. We did experiments on Aachen (day/night) and CMU Seasons datasets to show that our method retrieves more relevant database images compared to NetVLAD that leads to accurate query camera pose estimation.
 For each query, 20 images from the final shortlist produced by our method and NetVLAD were fed into a baseline localization pipeline, which uses a RANSAC PnP solver to register the query using 2D-3D matches (produced by performing 2D matching between the query and database images using our network and then utilizing known semi-dense point cloud for database images) and does hypothesis selection based on inlier count. 
 We report the proportion of correctly localized queries for the threshold ($5m,10^{\circ}$) in the following. Aachen day: 81.7\% (NetVLAD-Pitts), \textbf{84.7}\% (ours). Aachen night: 64.3\% (NetVLAD-Pitts), \textbf{68.4}\% (ours). CMU: 78.9\% (NetVLAD-Pitts), \textbf{89.1}\% (ours).
 Better accuracy in query camera pose estimation \emph{given the same localization pipeline and image matching method} shows that our approach retrieves higher quality database images compared to NetVLAD-Pitts (\cf~Tab.~\ref{tab:aachen_cmu_loc}). Our verification framework is generic and can be plugged in to other localization systems, such as HF-Net~\cite{sarlin2019coarse} and D2-Net~\cite{Dusmanu2019CVPR}. 



\begin{figure*}[t!]
 	\centering
 	\begin{subfigure}[t]{.99\textwidth}
 		\centering
 		\includegraphics[width=\textwidth]{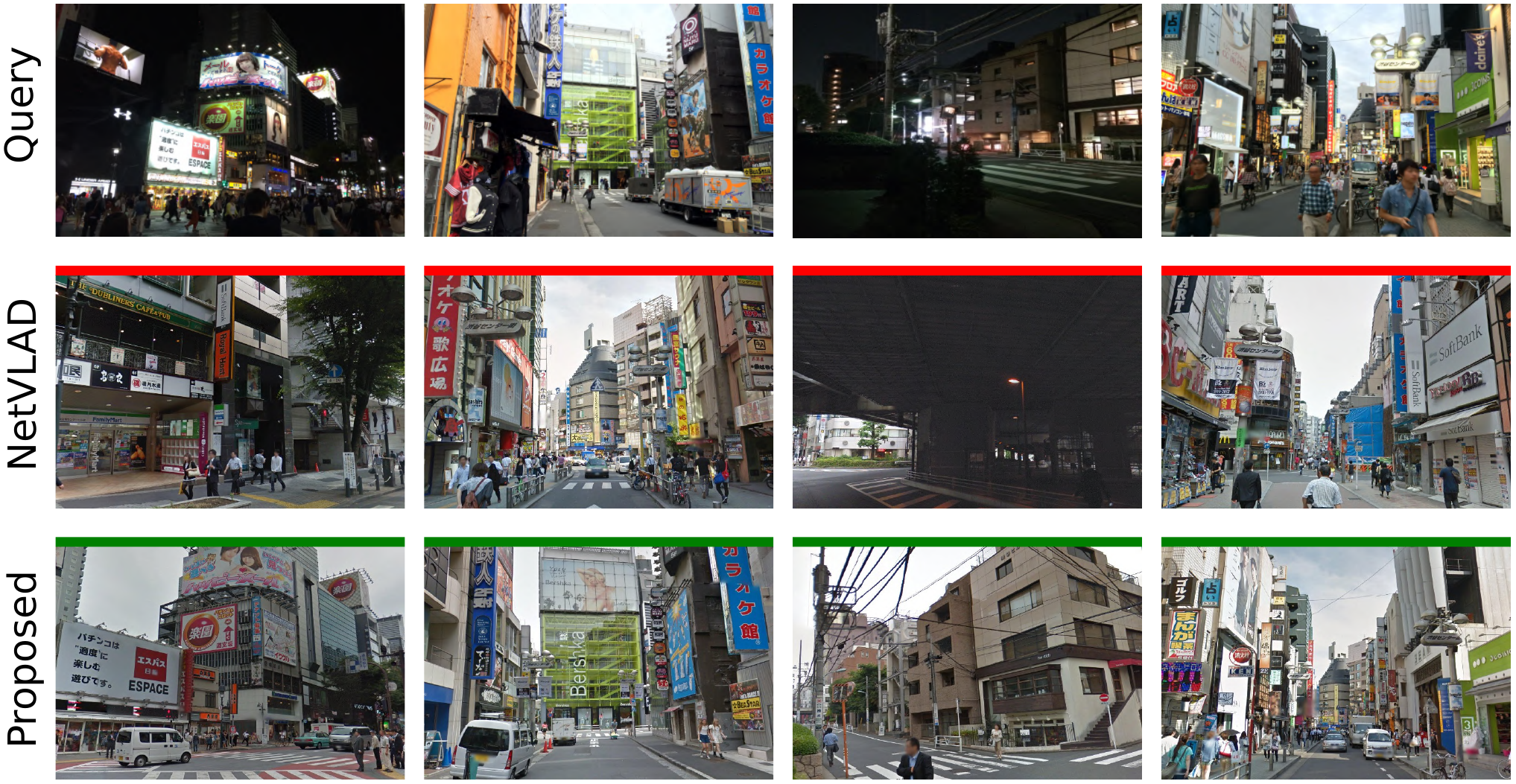}
 		\caption{Tokyo24/7}
 	\end{subfigure}
 	~
 	\begin{subfigure}[t]{.99\textwidth}
 		\centering
 		\includegraphics[width=\textwidth]{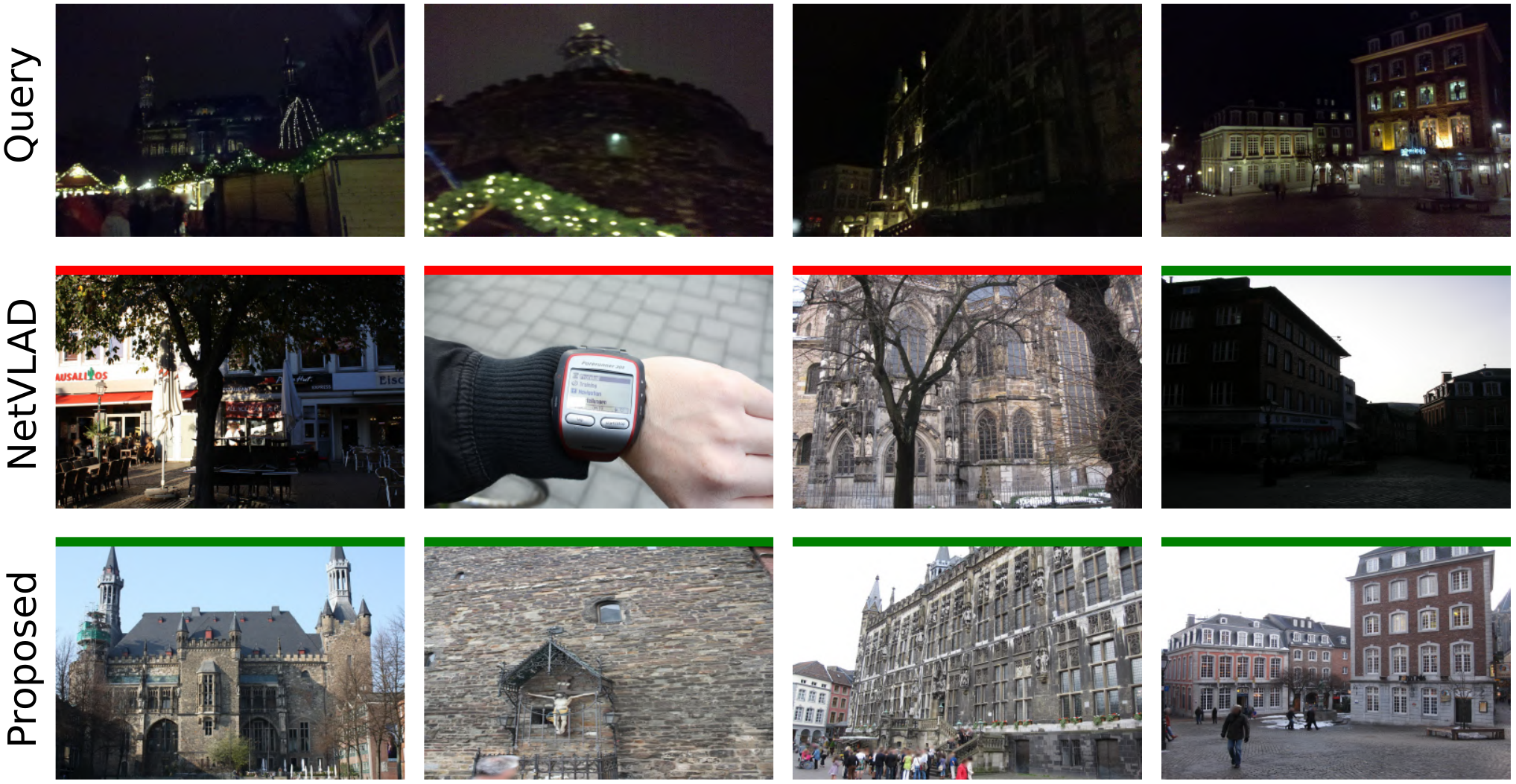}
 		\caption{Aachen Day-Night}\label{sfig:aachen_quality}
 	\end{subfigure}%
\caption{\textbf{Qualitative results} produced by NetVLAD~\cite{Netvlad} (rows 2 and 5) and the proposed method (rows 3 and 6) on two localization datasets: Tokyo24/7 and Aachen Day-Night. Each column corresponds to one test case: for each query (row 1 and 4) top-1 (Recall@1) nearest database image has been retrieved. The green and red strokes correspond to correct and incorrect retrieved images, respectively. The proposed approach can handle different illumination conditions (day/night) and significant viewpoint changes (the second column in Fig.~\ref{sfig:aachen_quality}). More examples presented in the supplementary.}\label{fig:quality_retrieval_tokyo_aachen}
\vspace{-2mm}
\end{figure*}

Qualitative image retrieval results on Tokyo 24/7 and Aachen Day-Night datasets are illustrated in Fig.~\ref{fig:quality_retrieval_tokyo_aachen}.


\begin{table}[t!]
\begin{center}
\resizebox{.43\textwidth}{!}{%
    \begin{tabular}{l| c c | c c c}
        \multirow{3}{*}{Methods} & \multicolumn{5}{|c}{Condition, $5m, 10^{\circ}$} \\
        & \multicolumn{2}{c|}{Aachen Day-Night} & \multicolumn{3}{c}{CMU-Seasons} \\
        & \multicolumn{1}{|c}{day} & \multicolumn{1}{c|}{night} & \multicolumn{1}{c}{urban} & \multicolumn{1}{c}{suburban} & \multicolumn{1}{c}{park}\\
    \hline
    \hline
    HF-Net~\cite{sarlin2019coarse} & 94.2 & 76.5 & 97.9 & 92.7 & 80.4\\
    D2-Net~\cite{Dusmanu2019CVPR} & 93.4 & 74.5 & - & - & -\\
    Active Search~\cite{ActiveSearch} & 96.6 & 43.9 & - & - & - \\
    NetVLAD-Pitts & 81.7 & 64.3 & 78.9 & 77.0 & 63.2 \\
    \hline
    Proposed & \textit{84.7} & \textit{68.4} & \textit{89.1} & \textit{77.1} & \textit{63.3} \\
    \hline
\end{tabular}
}
\end{center}
\vspace{-3mm}
\caption{Localization performance on the Aachen and CMU-Seasons datasets (higher is better). The best performance among \textit{image retrieval} based approaches is highlighted as \textit{italic}.}\label{tab:aachen_cmu_loc}
\vspace{-2mm}
\end{table}

\subsection{Ablation study}\label{ssec:ablation}

As the proposed UCMD decoder, $D_c$ is defined as a refinement function operating on the space of representation similarity, it should be invariant to the representations themselves. This allows us to replace the VGG-16 encoder with a much light weight encoder MobileNetv2 (MNetv2)~\cite{mobilenetv2} at test time without further re-training while keeping the same decoder trained on the features produced by VGG-16. This leads to a highly compact model. In practice, localization problem is most relevant in the context of mobile devices, thus the model compactness is crucial also at test time. This led to comparable performance on the challenging Tokyo24/7 dataset (\cf Tab.~\ref{tab:tokyo_loc_ablation}) and reduced the total number of network parameters from 8\textbf{M} (VGG16:$\sim$7\textbf{M}, UCMD:$\sim$0.9\textbf{M}) to 1\textbf{M} (MNetv2:$\sim$0.07\textbf{M}, UCMD:$\sim$0.9\textbf{M}). In the latter case UCMD provides notable memory savings compared to the original DGC-net (~10\textbf{M}). One feed-forward pass through the DGC-NC-UCMD-Net with MNetv2 encoder requires 60ms compared to 80ms with VGG16 encoder providing savings in computation time.

\begin{table}[t!]
\begin{center}

\resizebox{.42\textwidth}{!}{%
    \begin{tabular}{l|l l l}
        \multirow{2}{*}{Methods} & \multicolumn{3}{|c}{Recall} \\
        & \multicolumn{1}{|c}{r@1} & \multicolumn{1}{c}{r@5} & \multicolumn{1}{c}{r@10}\\
    \hline 
    Proposed $(S)$ Pitts & 71.43 & 82.54 & 85.08 \\
    Proposed $(S)$ (MNetv2 enc.) Pitts & 73.02 & 81.9 & 85.4 \\
    \hline
    Proposed $(S_F)$ Pitts & \textbf{77.14} & \textbf{84.44} & \textbf{86.67} \\
    Proposed $(S_F)$ (MNetv2 enc.) Pitts & 76.51 & 83.17 & 84.13 \\
\end{tabular}
}
\end{center}
\caption{\textbf{Ablation study}. Localization performance on the Tokyo24/7 dataset (MobileNetv2 decoder).}\label{tab:tokyo_loc_ablation}
\end{table}


\section{Conclusion}
We have presented novel methods for CNN based dense pixel to pixel correspondence learning and its application to geometric verification for image retrieval. In particular, we have proposed a compact but effective CNN model for dense pixel correspondence estimation using the universal correspondence map decoder block. Due to the universal nature of the decoder, we are able to obtain memory and computational savings at evaluation time.

In addition, we have integrated the matching layer in our model with neighborhood consensus~\cite{RoccoNeigh} which further enhances the matching performance. This modified dense correspondence model along with the proposed geometric similarity functions are then applied to improve the initial ranking of database images given by NetVLAD descriptor. We have evaluated our approach on three challenging city-scale localization datasets achieving state-of-the-art retrieval results.

\appendix
\section*{Appendix}\label{sec:appendix}
In this appendices we show additional qualitative and quantitative results of the proposed approach. In Sec.~\ref{sec:supp_ablation_retrieval} we provide an ablation study and analyze the influence of different design choices of our method to the localization performance. We demonstrate the benefits of the unified correspondence map decoder (UCMD) compared to the architecture with multiple decoders in Sec.~\ref{sec:supp_ucmd_benefits} 
Finally, qualitative localization and pixel correspondence estimation results are shown in Sec.~\ref{sec:supp_quality}.

\section{Additional Baselines}

In this work, we propose two similarity functions for geometric verification,~\ie:

\begin{equation}\label{eq:s_sim_app}
S = \frac{C}{I} \cdot exp\left(-\frac{W \cdot H}{C}\right),
\end{equation}

\begin{equation}\label{eq:final_sim_app}
    S_F = \overbrace{\log_{10}\left(S_L \cdot S\right)}^{R} \cdot \underbrace{10^{-G}}_{Q}
\end{equation}
\noindent where $I$ and $C$ is the number of inliers and cyclically consistent inliers between two $\left(W,H\right)$ images $\left(A, B\right)$, respectively; $S_L = \sum_{a} \left(f_{A}^a \cdot f_{B}^a\right) m^a$ is the local similarity between each hypercolumn ($f_{A}^a$ and $f_{B}^a$) of the NetVLAD~\cite{Netvlad} image descriptor at location $a$; $G$ is the global similarity value.

We compare our method with two baselines: \textit{i)} recently proposed geometric verification pipeline Inloc \cite{taira2018inloc}, and \textit{ii)} a neural network based method that learns the scoring functions, $S$ and $S_F$ given \{$C,I$\}, and $\{C,I,S_L,G\}$ as input. We present more details about the baselines next.

\noindent \textbf{Inloc.} Inloc is a indoor localization pipeline consisting of three primary stages: \textbf{i}) ranking of database images by measuring global representation similarity with a given query. The global representations are obtained from the image retrieval pipeline,~\eg NetVLAD~\cite{Netvlad}; \textbf{ii}) a shortlist of top ranked database images are re-ranked based on geometric verification using dense CNN descriptors. The dense descriptors are obtained from different layers of the NetVLAD pipeline followed by a coarse to fine matching using nearest-neighbor search. The geometric verification is done using a standard RANSAC based inlier count. The final score is the sum of global similarity and inlier count; \textbf{iii}) the top ranked geometrically verified database images are fed into a pose verification stage. The final stage first estimates candidate query poses~\wrt the current shortlisted database images. The estimated pose is then verified using view synthesis, a process requiring dense database depth maps.
Our proposed geometric verification pipeline is similar to Inloc components i) and ii). The pose verification stage requires depth maps which is not always available. Therefore, we evaluate Inloc pipeline until the geometric verification stage and report results in Tab.~\ref{tab:tokyo_loc_abl1}.

\noindent\textbf{Learnt similarity functions.} Since both Eq.~\ref{eq:s_sim_app} and Eq.~\ref{eq:final_sim_app} are hand-crafted we provide a FCNN-based model that can \textit{learn} the similarity function. More specifically, we experiment with two independent models (for $S$ and $S_F$) which can predict whether two images similar or not based on $C$, $I$, $S_L$, and $G$. Both models have similar architectures $FC(N,128)-ReLU-FC(128,128)-ReLU-FC(128,1)$, where the shorthand notation is used was the following: FC is a fully connected linear layer; $N$ is the number of input units (2  $\left\{C,I\right\}$ for $S$ and 4 $\left\{C,I,S_L,G\right\}$ for $S_F$, respectively). We refer to these models as $S$-FCNN and $S_F$-FCNN. Both models have been trained by minimizing binary cross-entropy loss function in a supervised manner.

\textbf{Results.} We now compare $S$ and $S_F$ with Inloc geometric verification pipeline on Tokyo247 dataset. Results demonstrate that our proposed function $S$ and $S_F$ outperform Inloc across all Recall rates as shown in Tab.~\ref{tab:tokyo_loc_base3}. We observed that for many query-database image pairs, Inloc fails to find any inliers. This can be attributed to significant clutter, illumination change (day-night) and occlusion in this challenging dataset. The learnt similarity functions $S$-FCNN and $S_F$-FCNN have very promising results and perform better than NetVLAD. In particular, $S$-FCNN has comparable performance to the proposed $S$. However, $S_F$-FCNN could not achieve any improvement compared to $S$-FCNN. We leave further analysis for future work.

\begin{table}[h!]
    \begin{subtable}[h]{0.49\textwidth}
        \centering
        \resizebox{.8\textwidth}{!}{%
    \begin{tabular}{l|l l l}
        \multirow{2}{*}{Methods} & \multicolumn{3}{|c}{Recall} \\
        & \multicolumn{1}{|c}{r@1} & \multicolumn{1}{c}{r@5} & \multicolumn{1}{c}{r@10}\\
    \hline
    \hline
    Inloc~\cite{taira2018inloc} & 62.54 & 67.62  & 70.48 \\
    NetVLAD-Pitts~\cite{Netvlad} & 61.27 & 73.02 & 78.73 \\
    DenseVLAD~\cite{Tokyo24/7} & 67.10 & 74.20 & 76.10 \\
    \hline
    $S$-FCNN                   & 67.94 & 81.90  & 85.08 \\
    $S_F$-FCNN              & 63.49 & 81.59  & 85.71 \\      
    \hline
    Proposed $(S)$ Pitts & 71.43 & 82.54 & 85.08 \\
    Proposed $(S_F)$ Pitts & \textbf{77.14} & \textbf{84.44} & \textbf{86.67} \\
    \hline
    $S_F$ + SuperPoint & \textit{83.17} & \textit{86.35} & \textit{87.94}
\end{tabular}
        }
    \caption{The proposed similarity functions $S$ and $S_F$ perform better strong baseline methods. An additional result after publication shows that combining our approach on SuperPoint keypoint locations gives improved retrieval results.}\label{tab:tokyo_loc_base3}    
    \end{subtable}
        
    \begin{subtable}[h]{.49\textwidth}
        \centering 
        \resizebox{.9\textwidth}{!}{%
            \begin{tabular}{l|l l l}
        \multirow{2}{*}{Methods} & \multicolumn{3}{|c}{Recall} \\
        & \multicolumn{1}{|c}{r@1} & \multicolumn{1}{c}{r@5} & \multicolumn{1}{c}{r@10}\\
    \hline
    \hline
    NetVLAD-Pitts~\cite{Netvlad} & 61.27 & 73.02 & 78.73 \\
    \hline
    $I$ (inliers)                &  56.83 & 78.41   & 83.81 \\
    $C$ (cyclically consistent inliers)      &  70.16 & 82.86   & 85.71 \\
    $C/I$                         &  64.76 & 82.54   & 85.71 \\           
        
    Proposed $(S)$ Pitts & 71.43 & 82.54 & 85.08 \\
    
    \hline
\end{tabular}
        }
        \vspace{-1mm}
        \caption{Localization performance on the Tokyo247 dataset (higher is better).}\label{tab:tokyo_loc_abl1}
    \end{subtable}
    
    \begin{subtable}[h]{.49\textwidth}
    \centering
    \resizebox{.8\textwidth}{!}{%
        \begin{tabular}{l|l l l}
        \multirow{2}{*}{Methods} & \multicolumn{3}{|c}{Recall} \\
        & \multicolumn{1}{|c}{r@1} & \multicolumn{1}{c}{r@5} & \multicolumn{1}{c}{r@10}\\
    \hline
    \hline
    NetVLAD-Pitts~\cite{Netvlad} & 61.27 & 73.02 & 78.73 \\
    \hline
    $\log_{10}({S_L*S})$        &  73.65 & 83.49   & 86.67 \\           
    $G$                         &  69.84 & 80.95   & 85.08 \\           
    
    Proposed $(S_F)$ Pitts & \textbf{77.14} & \textbf{84.44} & \textbf{86.67} \\
    \hline
\end{tabular}
        }
        \caption{Localization performance on the Tokyo247 dataset (higher is better).}\label{tab:tokyo_loc_abl2}
        \end{subtable}
        \vspace{-2mm}

  \begin{subtable}[h]{.49\textwidth}
    \centering
    \resizebox{.55\textwidth}{!}{%
     \begin{tabular}{|l|c|c|}\hline
        \diagbox[width=6em]{$R$}{$Q$} & $5/G$ & $10^{-G}$ \\ \hline
        $S_L$ & 77.78 & 81.34 \\ \hline
        $S_L * S$ & 82.04 & 83.70 \\ \hline
        $\log_{10}\left(S_L*S\right)$ & 85.37 & \textbf{85.94} \\ \hline     
    \end{tabular}
        }
        \caption{Localization performance (Recall@1) on the \textit{Pittsburgh} test dataset (higher is better). We analyze the performance of different $Q$ and $R$ of the original similarity function~(\ref{eq:final_sim_app}). The baseline, NetVLAD achieves 81.59 $\%$ Recall@1}\label{tab:pitts_loc_abl_rq}
        \end{subtable}
        \vspace{-2mm}
        \caption{\textbf{Ablation study.} We evaluate the proposed similarity functions $S$ and $S_F$ with different settings on Tokyo24/7 and Pittsburgh datasets.} \label{tbl:ablation_study}
\end{table}

\section{Ablation study}~\label{sec:supp_ablation_retrieval}


In this section we perform an ablation study on the proposed equations Eq.~\ref{eq:s_sim_app} and Eq.~\ref{eq:final_sim_app} for geometric verification. For Eq.~\ref{eq:s_sim_app}, we analyze the impact of each variable, {$C,I$} on retrieval performance on Tokyo247 dataset independently. The results are presented in Tab. \ref{tab:tokyo_loc_abl1}. 

\noindent \textbf{Results.} First we provide the ablation study for Eq.~\ref{eq:s_sim}. Results demonstrate that simple Inlier count performs worse than the baseline NetVLAD and our proposed $S$ at Recall@1. However, the retrieval performance improves over NetVLAD for Recall@5 and Recall@10. Cyclically consistent inliers $C$ outperform  NetVLAD across various Recall rates. Similarly, the ratio $C/I$ performs marginally better but it falls slightly behind of $C$ for Recall@1 (by about 6 $\%$). Both $C$ and $C/I$ perform on par with the proposed $S$ across Recall@5 and Recall@10. However, $S$ has a clear performance advantage over $C$ and $C/I$ for Recall@1 as shown in Tab. \ref{tab:tokyo_loc_abl1}.

Now, we perform an ablation study for Eq.~\ref{eq:final_sim_app}. As mentioned in the main manuscript, the proposed $S_F$ is used to re-rank the top 20 database images in the shortlist, $L$ as ranked by $S$. Here, we perform the final re-ranking using just the local descriptor similarity component, $\log_{10}(S_L * S)$, and global representation distance, $G$. Results in Tab. \ref{tab:tokyo_loc_abl2} demonstrate that re-ranking with $G$ decreases retrieval performance compared to the initial ranking by $S$. On the other hand, local descriptor similarity $S_L$ weighted by $S$ significantly improves over the baselines and initial ranking by $S$. However, the proposed combination of local and global representation similarity outperforms each individual component across all Recall rates.

The key idea here is to combine the similarity functions, $S_L$, $S$ and $G$. It is important to note that $S_L$ and $S$ are similarity functions, while $G$ is a distance function, hence, it is inversely proportional to global similarity. The inversely proportional functions, $R(S_L,S)$ and $Q(G)$ can be combined in many different ways. We present a few in Tab.~\ref{tab:pitts_loc_abl_rq}. The co-efficient (5 and 10) associated with $G$ in the columns of the Tab.~\ref{tab:pitts_loc_abl_rq} have  been obtained using a grid search over the range $\left(1, 10000\right)$ on Pittsburgh test dataset. In addition, we found $\hat{S}_F=R*Q$ performs clearly better than $\hat{S}_F$ = $R+Q$. Hence, we only present results for various $R$ and $Q$ for $\hat{S}_F=R*Q$ in Tab.~\ref{tab:pitts_loc_abl_rq}. The precise form of the combination of these similarity functions has been obtained based on validation experiments on test set of Pittsburgh dataset. Tab.~\ref{tab:pitts_loc_abl_rq} shows that various possible combinations give better performance than NetVLAD which achieves 81.59 at Recall@1. 
The weighting with structural similarity $S$ leads to a significant boost in retrieval performance. Such a form of weighting provides good balance requiring image pairs to have high local ($S_L$) and structural similarity ($S$). Among the various combinations, the proposed Eq.~\ref{eq:final_sim_app} achieves the best performance (highlighted bold in Tab. \ref{tab:pitts_loc_abl_rq}).

\begin{figure}[t!]
  \centering
    \includegraphics[width=.8\linewidth]{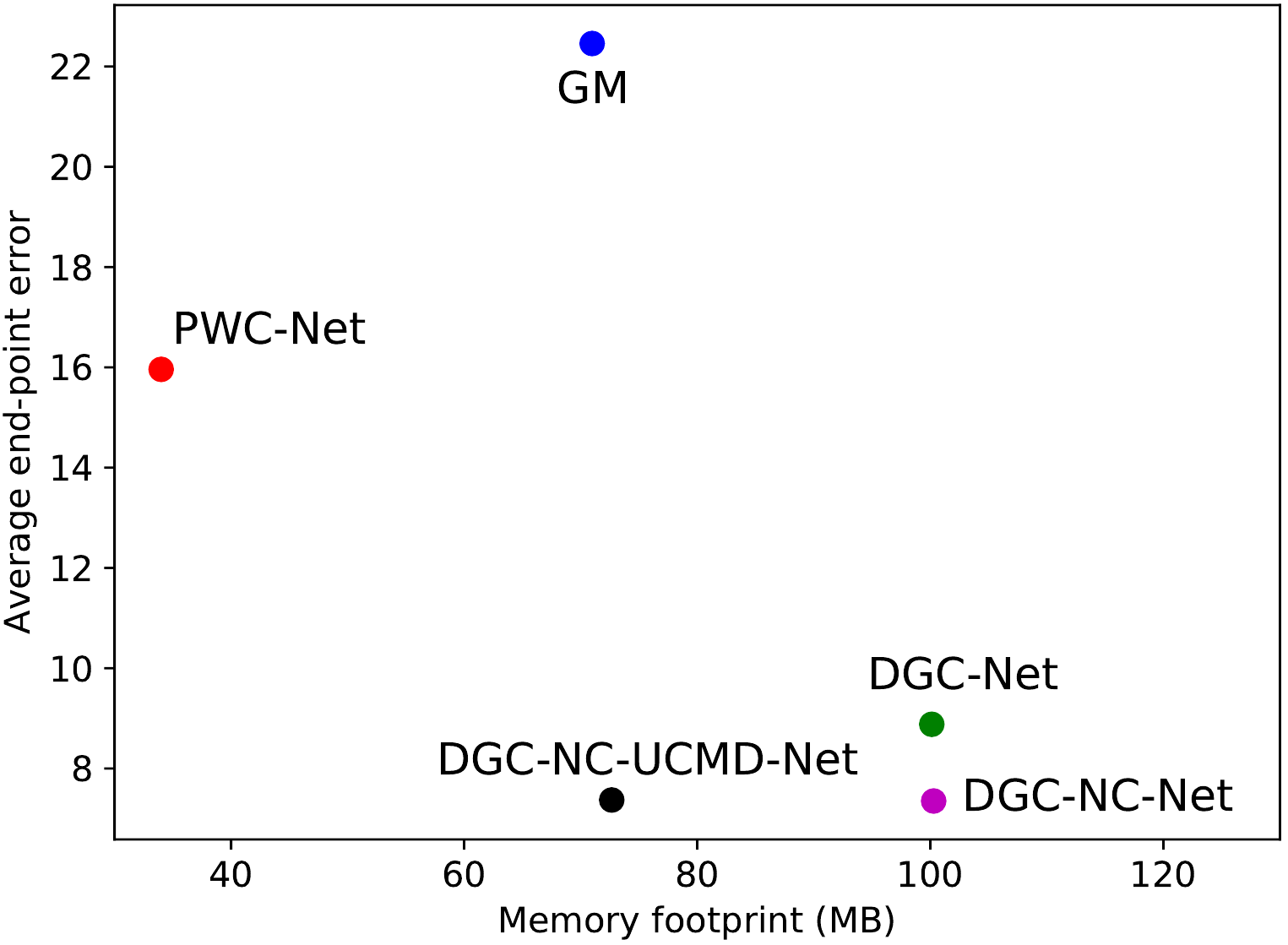}
  \caption{\textbf{AEPE averaged over all HPatches~\cite{HPatches} sequences versus memory footprint}. Accuracy of both proposed methods (DGC-NC-Net and DGC-NC-UCMD-Net) is about on par, however, UCMD allows to decrease memory footprint by~30\%.}
 \label{fig:f_memory_footpint}
 \vspace{-2mm}
\end{figure}

\section{The benefits of UCMD}~\label{sec:supp_ucmd_benefits}
As shown in the main manuscript, we propose the unified correspondence map decoder which leads to a compact but efficient architecture. In order to elaborate on the benefits of UCMD, here we report the average end point error averaged over all sequences of the HPatches~\cite{HPatches} dataset obtained by each strong baseline method (PWC-Net~\cite{PWC-Net}, geometric matching GM~\cite{Rocco17}, and DGC-Net~\cite{DGCnet}) and allocated GPU memory. The results are illustrated in Fig.~\ref{fig:f_memory_footpint}. In contrast to DGC-NC-Net with 5 separate decoders, the proposed UCMD can significantly decrease memory footprint (by 30\%) achieving comparable accuracy.

\begin{table}[ht]
\begin{center}
\resizebox{.5\textwidth}{!}{%
    \begin{tabular}{l c }
    \multicolumn{1}{c}{Model} & \multicolumn{1}{|c}{Number of learnable parameters}\\
        \hline
        \hline
        PWC-Net~\cite{PWC-Net} &  8 749 280 \\
        GM~\cite{Rocco17} &  3 271 576 \\
        DGC-Net~\cite{DGCnet} &  2 675 338 \\
        \hline
        Proposed (DGC-NC-Net) & 2 685 079\\
        Proposed (DGC-NC-UCMD-Net) & \textbf{940 561}\\
        \hline
    \end{tabular}
}
\end{center}
\caption{Number of learnable parameters of two proposed architectures and strong baseline methods.} \label{tbl:sm_bench_number_parameters}
\end{table}

The amount of memory allocated by GM~\cite{Rocco17}, DGC-Net~\cite{DGCnet}, DGC-NC-Net, and DGC-NC-UCMD-Net is higher compared to PWC-Net since all those models have used pre-trained VGG-16 network as encoder. Therefore, in addition to memory consumption, we compute the total number of learnable parameters of each model and provide the results in Tab.~\ref{tbl:sm_bench_number_parameters}. 

\section{Qualitative results}~\label{sec:supp_quality}
\begin{figure*}[t!]
  \centering
    \includegraphics[width=1\linewidth,height=20cm]{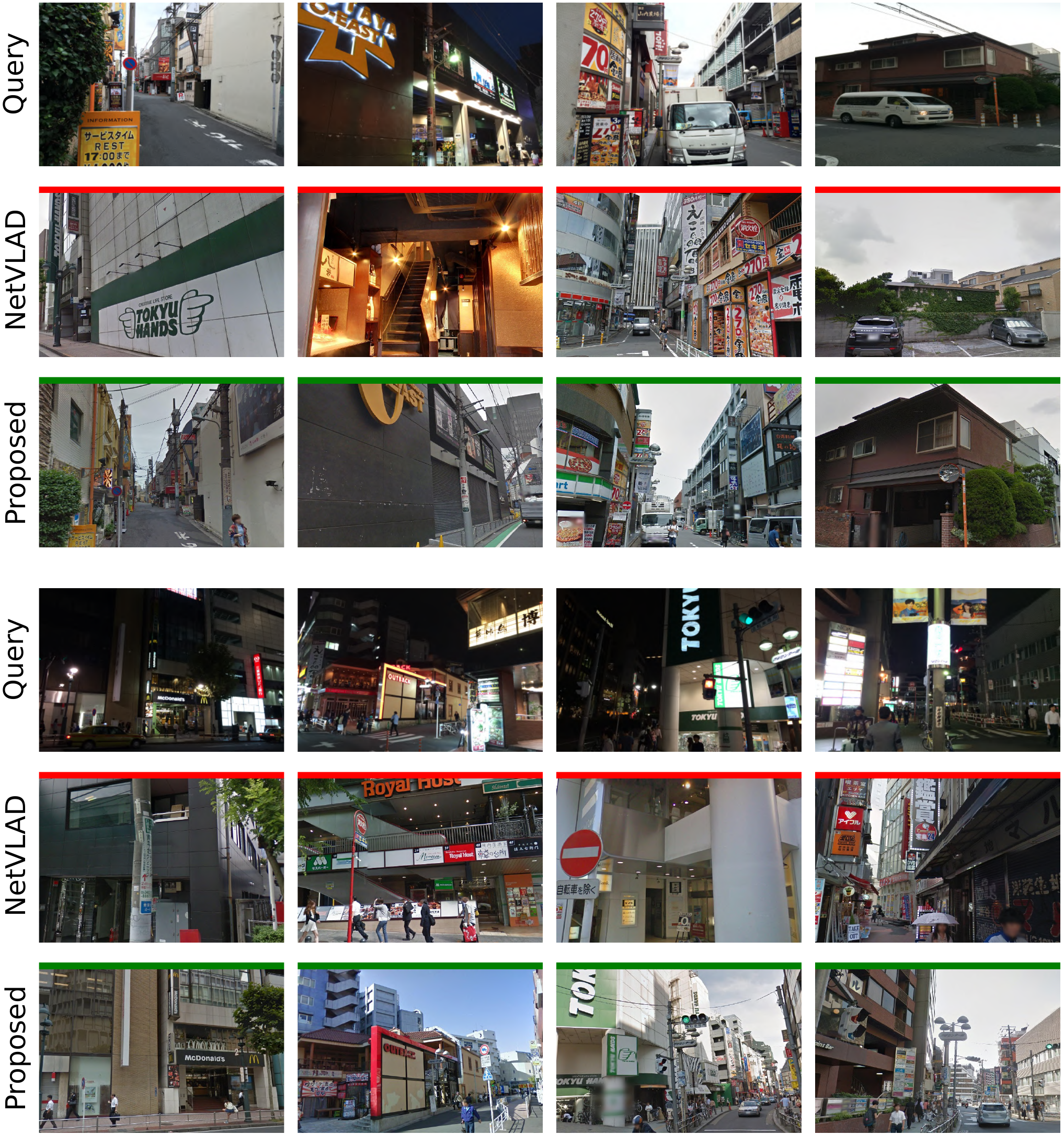}
  \caption{\textbf{Qualitative results} produced by NetVLAD~\cite{Netvlad} (rows 2 and 5) and the proposed method (rows 3 and 6) on Tokyo24/7~\cite{Tokyo24/7}. Each column corresponds to one test case: for each query (row 1 and 4) top-1 (Recall@1) nearest database image has been retrieved. The green and red strokes correspond to correct and incorrect retrieved images, respectively. The proposed approach can handle different illumination conditions (day/night) and significant viewpoint changes.}
 \label{fig:f_supp_qual_tokyo247}
\end{figure*}

\begin{figure*}[t!]
 	\centering
 	\begin{subfigure}[t]{.48\textwidth}
 		\centering
 		\includegraphics[width=.99\textwidth]{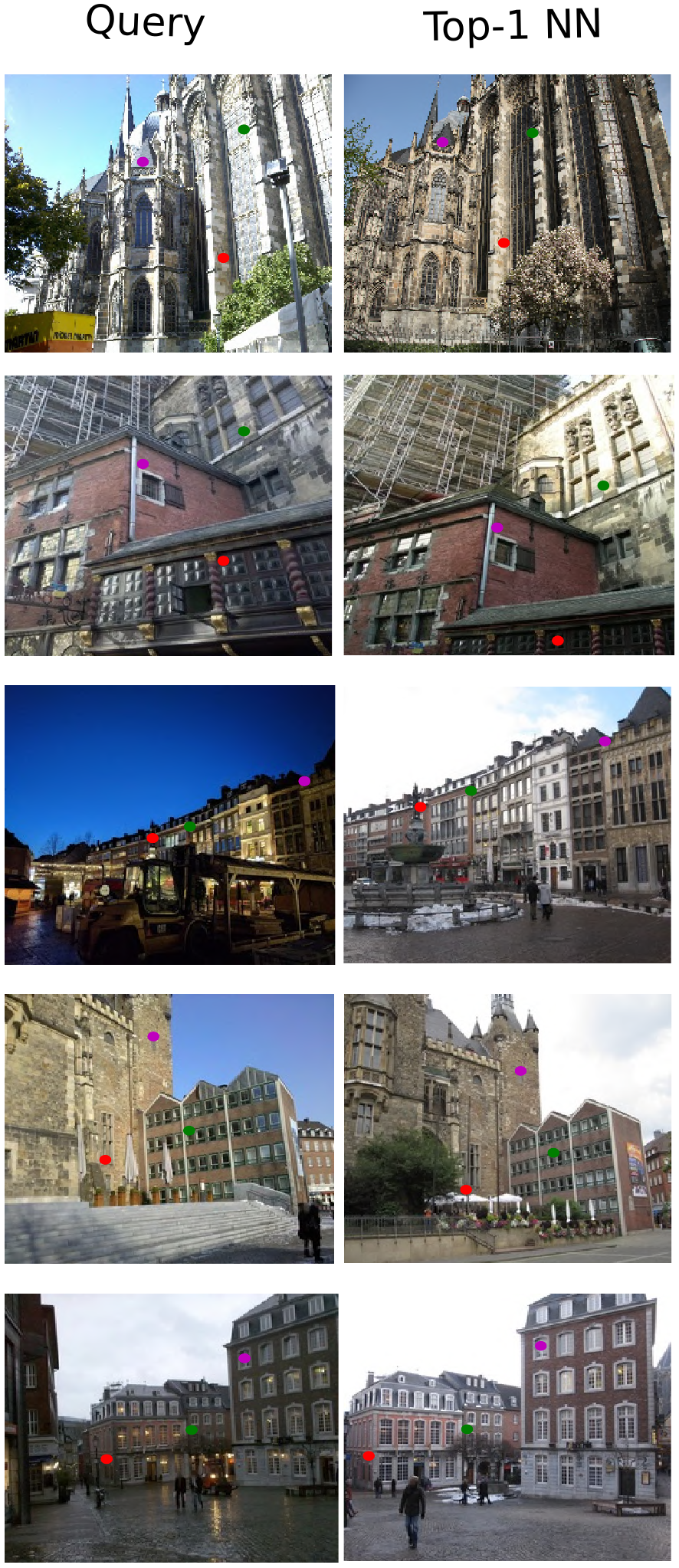}
 		\caption{Retrieval performance and pixel correspondences on Aachen Day-Night}\label{sfig:f_supp_qual_aachen_retrieval}
 	\end{subfigure}%
 	~
 	\begin{subfigure}[t]{.48\textwidth}
 		\centering
 		\includegraphics[width=.99\textwidth]{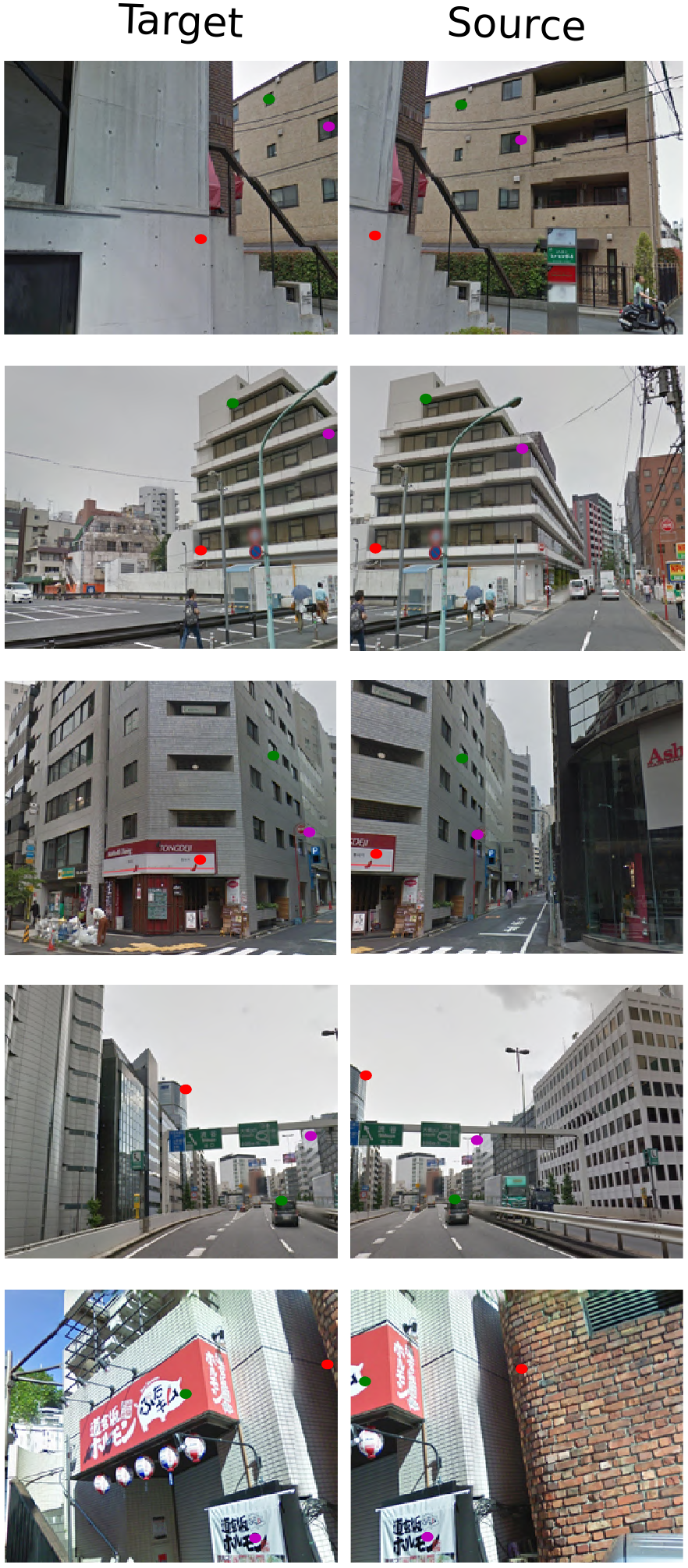}
 		\caption{Pixel correspondences on Tokyo24/7}\label{sfig:f_supp_qual_pixel_tokyo}
 	\end{subfigure}
\caption{\textbf{Qualitative image retrieval~\ref{sfig:f_supp_qual_aachen_retrieval} and dense pixel correspondence estimation results} produced by the proposed approach. We evaluate our approach on two challenging datasets: Tokyo24/7 and Aachen Day-Night. More image retrieval results are illustrated in Fig.~\ref{fig:f_supp_qual_tokyo247}. Each row of Fig.~\ref{sfig:f_supp_qual_pixel_tokyo} corresponds to one test case. Ground truth keypoints have been manually selected in the target image for visualization purposes and the corresponding locations in the source image are obtained by the proposed dense matching method. Keypoints of the same color are supposed to match each other.}\label{fig:f_supp_qual_pixel_tokyo}
\vspace{-2pt}
\end{figure*}

\textbf{Localization (image retrieval) performance.} Fig.~\ref{fig:f_supp_qual_tokyo247} reports an additional set of results obtained for the Tokyo24/7 dataset. Namely, it includes top-1 Nearest Neighbour (Recall@1 metric) obtained by NetVLAD~\cite{Netvlad} and our approach, respectively, for a given query. It clearly shows the proposed method improves retrieval results compared to NetVLAD and can cope with major changes in appearance (illumination changes in the scene) between the database and query images. Qualitative image retrieval results on Aachen Day-Night~\cite{AachenDataset} are illustrated in~Fig.~\ref{sfig:f_supp_qual_aachen_retrieval}.

\textbf{Dense pixel correspondences} are presented in Fig.~\ref{fig:f_supp_qual_pixel_tokyo}. Each row shows one test pair from the Aachen Day-Night and Tokyo24/7 datasets, respectively. Ground truth matching keypoints are illustrated in different colors and have been used only for pixel correspondence evaluation. Keypoints of the same color are supposed to match each other. We manually indicated 3 keypoints in the target image for visualization purposes and the corresponding locations in the source image have been obtained by the proposed automatic dense matching approach. That is, given an input image pair (source and target images), our method predicts the correspondence map which is then used to obtain the location of keypoints. The results demonstrate that the proposed method can handle such challenging cases as different illumination (day/night) conditions, occlusions, and significant viewpoint changes producing accurate pixel correspondences.

\section{Limitations and future directions}
We have demonstrated that the proposed method can localize queries under challenging conditions but it fails for very large viewpoint change (\eg $180^o$ rotation while observing the same place) and significant scale change. In addition, it would be interesting to propose an end-to-end semi-supervised approach which can efficiently learn similarity functions.

The proposed similarity scores are computed at the cyclically consistent pixel locations which can be redundant due to the dense prediction by DGC-Net. A preliminary experiment showed that applying the proposed similarity functions at unique keypoint locations given by SuperPoint~\cite{superpoint} results in improved retrieval performance as shown in Tab.~\ref{tab:tokyo_loc_base3}. However, this comes at an additional cost of evaluating the keypoint detector. A possible future direction can be to train a joint keypoint detector and matching network.

{\small
\bibliographystyle{ieee}
\bibliography{egbib}
}

\end{document}